\documentclass[12pt]{article}

\usepackage[authoryear]{natbib}

\usepackage{array}
\usepackage[utf8]{inputenc} 
\usepackage[T1]{fontenc}    
\usepackage{hyperref}       
\usepackage{url}            
\usepackage{booktabs}       
\usepackage{amsfonts}       
\usepackage{nicefrac}       
\usepackage{microtype}      
\usepackage{bm}
\usepackage{graphicx}
\usepackage{placeins}
\usepackage{caption}
\usepackage{subcaption}
\usepackage{amsmath,amsfonts,amssymb}

\DeclareMathOperator{\cdf}{cdf}

\DeclareMathOperator*{\argmax}{arg\,max}
\DeclareMathOperator*{\onehot}{one\_hot} 

\usepackage[labelfont=bf,labelsep=period,justification=raggedright]{caption}
\usepackage{subcaption}

\begin{document}

\begin{flushleft}
{\Large
\textbf{A learning framework for winner-take-all networks with stochastic synapses}
}
\\
Hesham Mostafa and Gert Cauwenberghs
\\
Institute of Neural Computation \\
University of California San Diego\\
\texttt{\{hmmostafa,gert\}@ucsd.edu}\\
\end{flushleft}

%

\begin{abstract}
  Many recent generative models make use of neural networks to transform the probability distribution of a simple low-dimensional noise process into the complex distribution of the data. This raises the question of whether biological networks operate along similar principles to implement a probabilistic model of the environment through transformations of intrinsic noise processes. The intrinsic neural and synaptic noise processes in biological networks, however, are quite different from the noise processes used in current abstract generative networks. This, together with the discrete nature of spikes and local circuit interactions among the neurons, raises several difficulties when using recent generative modeling frameworks to train biologically motivated models. In this paper, we show that a biologically motivated model based on multi-layer winner-take-all (WTA) circuits and stochastic synapses admits an approximate analytical description. This allows us to use the proposed networks in a variational learning setting where stochastic backpropagation is used to optimize a lower bound on the data log likelihood, thereby learning a generative model of the data. We illustrate the generality of the proposed networks and learning technique by using them in a structured output prediction task, and in a semi-supervised learning task. Our results extend the domain of application of modern stochastic network architectures to networks where synaptic transmission failure is the principal noise mechanism.




\end{abstract}

\section{Introduction}
The hypothesis that sensory perception is a process of active inference is key to explaining various perceptual phenomena~\citep{Gregory80,Lee_Mumford03}. One form of this hypothesis is that the brain maintains a probabilistic model of the environment which it then uses to infer latent causes and to fill missing or ambiguous details in the noisy sensory input it receives~\citep{Friston03}. Traditionally, there has been a clear distinction between probabilistic modeling approaches developed based on practical considerations~\citep{Ackley_etal85,Dayan_etal95,Kingma_Welling13,Goodfellow14}, and probabilistic models developed as mechanistic explanations of how the brain represents and manipulates probability distributions~\citep{Deneve08,Ma_etal06,Mostafa_etal15}. While the later models are more biologically relevant, they lack the scalability and power of the former. Finding a middle ground between these two types of models could prove useful in two ways: insights gained from developing practical, large-scale, and biologically-motivated generative models could shed some light on how the brain is able to model complex high-dimensional distributions, and the constraints imposed by the biological substrate could inform the development of more computationally efficient types of generative models.

One example of an attempt to find such a middle ground used stochastic spiking networks to implement and sample from Boltzmann machines~\citep{Buesing_etal11}. This model was developed further through the use of more realistic neuronal dynamics~\citep{Neftci_etal14} and stochastic synapses~\citep{Neftci_etal16,Al-Shedivat_etal16}. Establishing a link between Boltzmann machines and the dynamics of biologically realistic networks, however, is difficult due to the need for a symmetric synaptic connectivity matrix. Moreover, generating samples from the Boltzmann distribution embodied by the spiking network (either unconditional samples or samples from the posterior distribution) requires running the network for several steps in order to approach the equilibrium distribution. For highly multi-modal distributions, significant number of steps might be needed for the Markov chain to mix, making it computationally expensive to sample from the network.

Restricted Boltzmann machines were among the first large-scale, effectively trainable generative models~\citep{Ackley_etal85,Hinton02}. Recently, however, effective variational training methods have been developed for training generative architectures with a feed-forward hierarchical structure of latent variables~\citep{Kingma_Welling13,Rezende_etal14} where generative samples can be obtained in a single pass through the network. Recent variational methods rely on having an analytically tractable distribution over the variables in one layer conditioned on their parent variables in the previous layer. For continuous latent variables, the distributions are typically Gaussians whose mean and variance are functions of the parent variables, while categorical/discrete variables typically follow a softmax distribution. Such distributions do not map naturally onto the dynamics of biologically-realistic networks. Moreover, they are not computationally cheap as they involve multiplications and exponentiation operations.



The main contribution of this paper is the development of an analytically tractable approximation of the probability distribution over possible network states in multi-layer networks of winner-take-all (WTA) circuits, where neurons in different WTA circuits are connected using stochastic synapses. Since the state of each WTA is a discrete variable (the identity of the winner), and we use samples from these discrete variables/WTAs to approximate various intractable expectations, we need to be able to backpropagate error information through these stochastic samples. The approximate expression for the probability distribution over network states that we develop  allows us to make use of the recently-introduced Gumbel-softmax approximate reparameterization of discrete distributions~\citep{Jang_etal17,Maddison_etal16} to enable backpropagation through stochastic network samples. We thus obtain a general learning framework for WTA networks with stochastic synapses that can be applied to a wide range of learning problems.

In order to obtain an analytically tractable approximation of the probability distribution over possible network states, it was necessary to use an abstract network model with no real temporal dynamics. This multi-layer abstract network of WTAs is evaluated layer-by-layer in the typical manner used to evaluate multi-layer artificial neural networks, and it can be trained using standard machine learning packages. We show that the parameters of this abstract network model can be directly mapped to the parameters of a network of leaky integrate and fire (LIF) neurons with stochastic conductance-based synapses. To our knowledge, this is the first general training method for feedforward non-rate-based spiking networks with stochastic synapses, allowing us to obtain spiking network versions of common stochastic architectures such as variational auto-encoders. This opens the way for implementing these stochastic architecture on power-efficient asynchronous neuromorphic chips~\cite{Qiao_etal15,Furber_etal14,Benjamin_etal14,Park_etal14}, as well as investigating their behavior on a biologically-realistic neural substrate. The feedforward structure sidesteps the biologically unrealistic requirement of symmetric weights found in spiking neural implementations of Boltzmann machines~\citep{Buesing_etal11}. The feedforward structure also allows very fast sampling with no need to run a long Markov chain to obtain unbiased samples. The training method we used is an off-line training method based on backpropagation, however. Further work is needed to develop a more biologically-motivated learning method, in the spirit of the learning method in ref.~\cite{Mostafa_etal17a}, that learns online and changes synaptic weights based only on information in the pre- and post-synaptic neurons.

We first use the proposed networks to solve a structured output prediction task in order to illustrate the soundness of our network approximation and of the training method. We then apply the proposed learning framework to learning a generative model of the MNIST dataset using variational methods. Following recent trends in variational methods, we use a stochastic neural network to implement the variational posterior. The network we use to approximate the posterior distribution over the latent variables is also based on WTA circuits with stochastic synapses. Both the generative/decoder branch and the inference/encoder branch of the network are thus composed solely of discrete valued WTA circuits. To illustrate the generality of the proposed networks, we also use them in a configuration that is inspired by ladder networks~\citep{Rasmus_etal15} to solve a semi-supervised learning task.

\section{Model description}
\label{sec:model}
\begin{figure}[t]
  \centering
  \begin{subfigure}[b]{0.65\textwidth}
    \includegraphics[width=\textwidth]{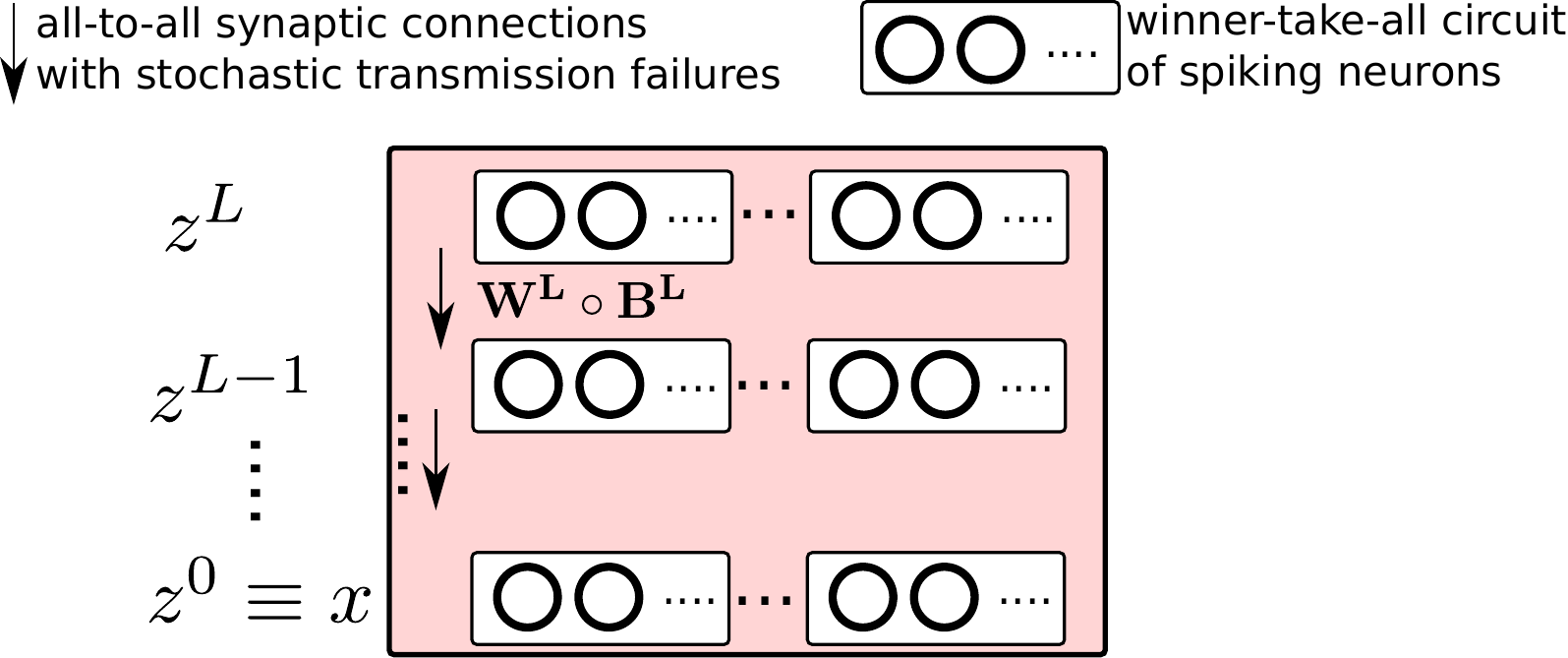}
    \subcaption{}
    \label{fig:model_fc}
  \end{subfigure}
  \quad
  \begin{subfigure}[b]{0.23\textwidth}
    \includegraphics[width=\textwidth]{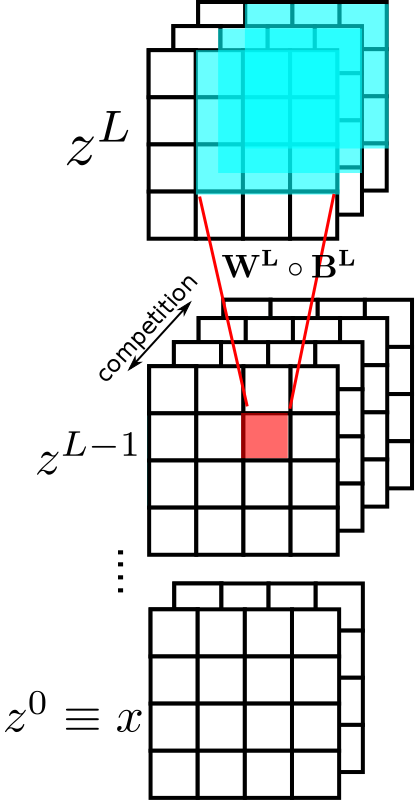}
    \subcaption{}
    \label{fig:model_conv}
  \end{subfigure}
    \caption{Multi-layer networks of WTA circuits. (\subref{fig:model_fc}) Network with fully connected layers of latent variables where each neuron projects to all neurons in the subsequent layer through connections that fail stochastically. (\subref{fig:model_conv}) Network with convolutional layers where each layer is composed of a number of feature maps. Competition is across the feature maps dimension, i.e, all neurons at the same spatial position in one layer are part of the same WTA. At each spatial position, only one neuron can win the competition and generate a spike. As in standard convolutional layers, each neuron receives input from a local spatial receptive field and the receptive field projection weights are tied across the spatial positions. The incidences of synaptic failure are not tied, however, and they are independent for each connection.}
    \label{fig:model}

\end{figure}

We investigate multi-layer networks of WTA circuits which have the general structure shown in Fig.~\ref{fig:model}. The network in Fig.~\ref{fig:model} has $L$ layers where each layer can have a different number of WTA circuits. WTA circuits in the same layer all have the same number of neurons. This number may be different in different layers.
Exactly one neuron in a WTA circuit spikes during one pass through the network and this is the neuron receiving the largest input among the WTA neurons. Neurons are connected using stochastic synapses where each synapse has an independent probability to fail to transmit a spike.

The inter-layer connection pattern can be all-to-all as in Fig.~\ref{fig:model_fc} or have a convolutional structure as in Fig.~\ref{fig:model_conv}. The convolutional connection pattern is the same as in standard convolutional networks with tied weights at the different spatial positions. The synaptic failure incidences are not tied, however; two connections having the same tied weight fail independently. In a layer receiving all-to-all connections, the neurons are arbitrarily grouped into WTA circuits. In a convolutional layer, a WTA circuit is formed by all neurons having the same spatial position, i.e, competition is across the feature maps dimension. Thus, at each spatial position, a spike is produced by only the most strongly activated feature map at that position. All convolutions are carried out using a stride of 1 and use zero padding to produce output feature maps having the same spatial dimensions as the input feature maps. For simplicity, the notation in the rest of the paper assumes all-to-all inter-layer connections described by a 2D weight matrix. This notation easily accommodates the convolutional connections case if parts of the 2D weight matrix are assumed to be fixed at zero to reflect the spatial locality of the receptive fields. 

Let ${\bf t^l}$ be the vector representing the total input to the neurons in layer $l$ and ${\bf z^l}$ a binary $0/1$ vector indicating whether each neuron in layer $l$ has spiked (1) or not (0).
Let $U^l(i)$ be a function that maps a neuron index $i$ in layer $l$ to the set of indices of all the neurons in the same WTA. For a convolutional layer, that would be all the neurons at the same spatial position.
${\bf t^l}$ and ${\bf z^l}$ are then given by

\begin{subequations}
  \begin{align}
  {\bf t^l} = ({\bf W^{l+1}} \circ {\bf B^{l+1}}){\bf z^{l+1}} &\quad\quad\quad l = 0,\ldots,L-1\\
  z_i^l = \begin{cases}
   1 & \text{if \quad $t_i^l = \max\limits_{j\in U^l(i)}t_j^l$}   \\
   0       & \text{otherwise}
   \end{cases} &\quad\quad\quad l = 0,\ldots,L-1,
  \end{align}
\end {subequations}
where $z_i^l$, $t_i^l$ are the $i^{th}$ elements in vectors ${\bf z^l}$ and ${\bf t^l}$ respectively. ${\bf W^l}$ is the weight matrix from layer $l$ to layer $l-1$ and ${\bf B^l}$ is a matrix of independent 0/1 Bernoulli random variables which is multiplied element-wise with ${\bf W^l}$. Each element in ${\bf B^l}$ can have a different probability of being $1$. These probabilities are given by the matrices ${\bf Z^l} = {\mathbb E}\left[{\bf B^l}\right]$. Each connection thus has an independent probability to fail to transmit a spike. The lowest layer is the data layer, $z^0\equiv x$. Note that network evaluation proceeds layer by layer and is thus an abstraction of the behavior of a dynamical spiking network.

We assume each neuron in the top layer has an equal probability to win the competition in its WTA circuit. This prior on the top layer activity, together with the network weights and synaptic transmission failure probabilities implicitly define a probability distribution over the spike configurations generated by the network. This distribution can be written as
\begin{equation}
\label{eq:distro}
\log\left(P_{\theta}({\bf z^L},{\bf z^{L-1}},\ldots ,{\bf z^1},{\bf x})\right) = \log\left( P({\bf z^L})\right)+\log\left( P_{\theta}({\bf x} \mid {\bf z^1})\right)+\sum_{l=1}^{L-1} \log\left(P_{\theta}({\bf z^{l}} \mid {\bf z^{l+1}})\right),
\end{equation}
where $\theta$ is the collection of all the network parameters (weights and transmission failure probabilities), and $P({\bf z^L})$ is the prior distribution over the top layer activity. 


It is easy to generate samples from the distribution given in Eq.~\ref{eq:distro} by sampling the top-layer prior and the failure probabilities of all the connections, then executing one pass through the network. In order to efficiently train these networks to optimize various expectations over the spike patterns generated by the network, we need to be able to quickly evaluate the probability of particular spike patterns, i.e, we need an explicit expression for the distribution in Eq.~\ref{eq:distro}. This means we need an explicit expression for the log conditional distributions $\log\left(P_{\theta}({\bf z^{l}} \mid {\bf z^{l+1}})\right)$. We derive this expression in the next subsection.

\subsection{Deriving the conditional distribution of the latent variables}
\begin{figure}[t]
  \centering
  \includegraphics[width=0.7\textwidth]{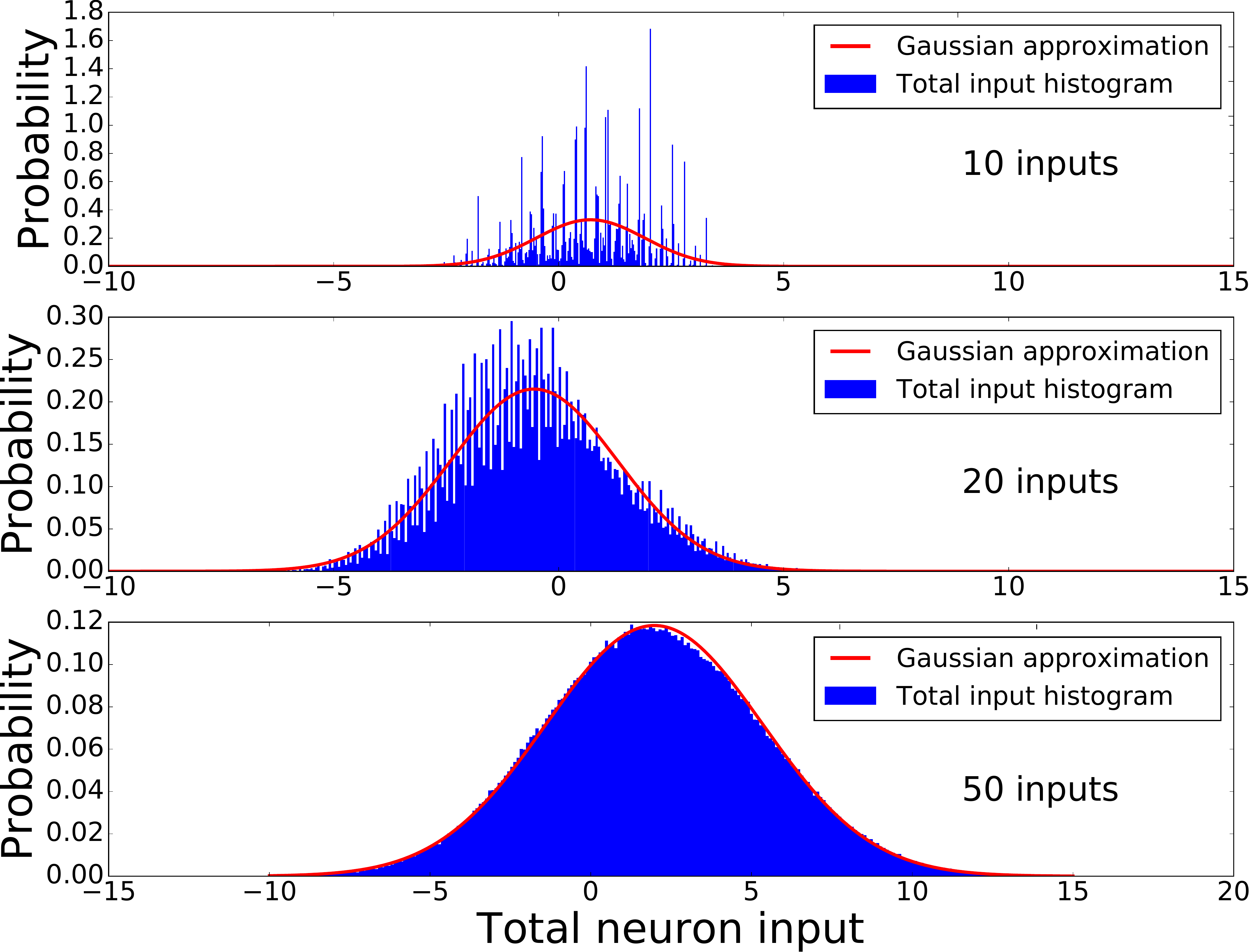}

\caption{Quality of the Gaussian approximation of the neuron's input distribution. We considered a neuron with 10, 20, and 50 inputs. The inputs are all 1s. The input connection weights were drawn randomly from a standard normal distribution. Each connection has an independent probability to fail which is drawn uniformly at random from $[0,1]$. After fixing the connection weights and failure probabilities, we generate 500000 samples for each of the three input sizes to construct the total input histogram.}
  \label{fig:bernoulli_fidelity}
\end{figure}

We consider a single WTA with $C$ neurons. The WTA has $C$ possible outputs where each output corresponds to a different neuron winning the competition and generating a spike. A neuron wins the competition in a WTA if it receives the largest input among the WTA neurons. Let $t_i$ be the input to the $i^{th}$ neuron of the WTA and ${\bf z^{in}}$ the binary vector representing the input spike activity from the preceding layer. $t_i$ and its first two moments are given by
\begin{align}
  t_i &= ({\bf w_i} \circ {\bf b_i})^T {\bf z^{in}} \label{eq:mean_variance}\\
  \mu( t_i) &= {\mathbb E}\left[t_i\right] = ({\bf w_i} \circ {\bm \zeta_i})^T{\bf z^{in}} \label{eq:mean}\\
  \sigma^2( t_i) &= {\mathbb E}\left[(t_i - \mu( t_i))^2\right] = ({\bf w_i}^2 \circ {\bm \zeta_i} \circ (1-{\bm \zeta_i}))^T{\bf z^{in}}, \label{eq:variance}
\end{align}
where ${\bf w_i}$ is the input weight vector for neuron $i$ in the WTA. ${\bf b_i}$ is a vector of Bernoulli random variables which model stochastic transmission failures and whose mean is ${\bm \zeta_i} = {\mathbb E}\left[{\bf b_i}\right]$. ${\bf b_i}$ is multiplied element-wise by ${\bf w_i}$. The mean and variance of $t_i$ are given by Eqs.~\ref{eq:mean} and~\ref{eq:variance} respectively. We make use of the central limit theorem to approximate the probability distribution of $t_i$ (which is the sum of many independent random variables) by a Gaussian having the same mean and variance. This approximation is quite accurate when the total number of non-zero inputs to the neuron is large. This is the number of non-zero entries in ${\bf z^{in}}$ which corresponds to the number of WTAs in the preceding layer. Figure~\ref{fig:bernoulli_fidelity} illustrates how the number of non-zero inputs affects the quality of the Gaussian approximation. For 10 and 20 inputs, the discrete nature of the neuron's input distribution is evident as there are only $2^{10}$ and $2^{20}$ possible inputs, respectively, corresponding to the possible configurations of synaptic transmission failure. For 50 inputs, the input distribution becomes much smoother and the Gaussian approximation becomes more accurate.

We first consider a WTA with two neurons whose total inputs are $t_1$ and $t_2$. The probability that $t_1 > t_2$ is the probability that $t_1$ takes a particular value multiplied by the probability that $t_2$ takes a smaller value, averaged across all values. Alternatively, the probability that $t_1>t_2$ is the probability that $r = t_1 - t_2 > 0$. Following the Gaussian approximation of $t_1$ and $t_2$, $r$ is also a Gaussian with mean $ \mu( t_1) - \mu( t_2)$ and variance $\sigma^2( t_1) + \sigma^2( t_2)$. The two equivalent ways of formulating $P(t_1 > t_2)$ are thus
\begin{align}
  P(t_1 > t_2) &= \int_{-\infty}^{\infty}\phi\left(x;\mu(t_1),\sigma(t_1)\right)\cdf\left(x;\mu(t_2),\sigma(t_2)\right)dx \label{eq:tractable_complicated}  \\ 
  P(t_1 > t_2) &= P(r > 0) = \cdf\left(\frac{\mu( t_1) - \mu( t_2)}{\sqrt{\sigma^2(t_1) + \sigma^2(t_2)}};0,1\right) \label{eq:tractable_simple} \\
  \phi\left(x;\mu,\sigma\right) &= \frac{1}{\sigma\sqrt{2\pi}} e^{-\frac{(x - \mu)^2}{2\sigma^2}}  \nonumber \\
  \cdf\left(x;\mu,\sigma\right) &= \frac{1}{\sigma\sqrt{2\pi}} \int_{-\infty}^{x}e^{-\frac{(y - \mu)^2}{2\sigma^2}} dy.\nonumber
\end{align}

\begin{figure}[t]
  \centering
  \includegraphics[width=0.7\textwidth]{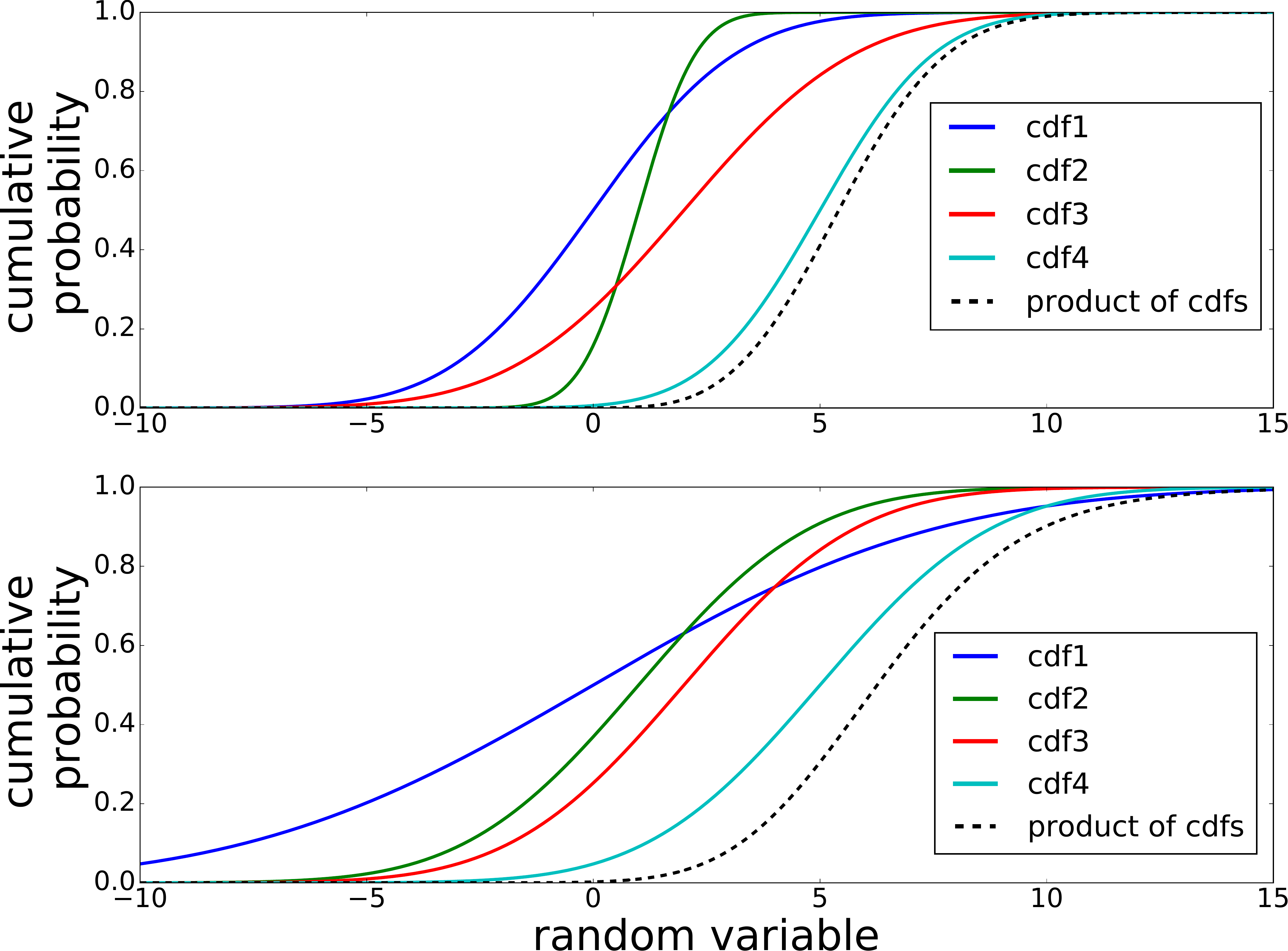}

  \caption{Approximating the product of four cumulative density functions (dashed black line) by one of the cumulative density functions making up the product. In the top plot, this product is approximated reasonably well by $\cdf4$. In the bottom plot, the approximation by $\cdf4$ is worse due to the larger variance of some of the cumulative density functions.}
  \label{fig:cdf_approx}

\end{figure}

For a WTA with $C$ neurons whose total inputs are $t_1,\ldots\,t_C$, and following the same reasoning as Eq.~\ref{eq:tractable_complicated}, the probability that neuron $i$ receives the largest input is
\begin{align}
  \label{eq:intractable}
  p_i &= P(\bigwedge\limits_{\substack { j\neq i}} t_i > t_j) = \int_{-\infty}^{\infty}\phi\left(x;\mu(t_i),\sigma(t_i)\right)\prod\limits_{\substack {j\neq i}} \cdf\left(x;\mu(t_j),\sigma(t_j)\right) dx.
\end{align}
The integration in Eq.~\ref{eq:intractable} is analytically intractable for $C > 2$. For $C=2$, $p_i$ reduces to the expressions in Eqs.~\ref{eq:tractable_complicated} and~\ref{eq:tractable_simple} . To approximate the integration in Eq.~\ref{eq:intractable}, we approximate the product of the $\cdf$ functions by one of the $\cdf$ functions making up the product. That will usually be the $\cdf$ of the Gaussian having the largest mean. This approximation is illustrated in Fig.~\ref{fig:cdf_approx}. The quality of the approximation improves when the means of the Gaussians are well separated and the Gaussians have low variance. This approximation is equivalent to approximating Eq.~\ref{eq:intractable} as the minimum probability that neuron $i$ wins a pairwise competition (governed by Eqs.~\ref{eq:tractable_complicated} and~\ref{eq:tractable_simple}) with another neuron in the WTA:
\begin{align}
  \hat{p_i} & = \min\limits_{\substack{j\neq i}}\left\{\int_{-\infty}^{\infty}\phi\left(x,\mu(t_i),\sigma(t_i)\right)\cdf\left(x;\mu(t_j),\sigma(t_j)\right) dx\right\} \label{eq:approx_integ}\\
  & = \min\limits_{\substack{j\neq i}}\left\{ \cdf\left(\frac{\mu( t_i) - \mu( t_j)}{\sqrt{\sigma^2(t_j) + \sigma^2(t_i)}};0,1\right)\right\} \label{eq:p_unnormalized}\\
  \tilde{p_i} & = \frac{\hat{p_i}}{\sum\limits_j \hat{p_i}}. \label{eq:p_normalized}
\end{align}
In other words, we approximate the probability that a neuron in a WTA wins the competition (receives the largest input) by the probability that it wins the competition against one other neuron, where this other neuron is selected to be the neuron which has the highest probability to receive a larger input than the first neuron. This approximation works in practice because winning the competition against the neuron that has the highest mean input implies winning the competition against other neurons as well with a high probability. This approximation yields an unnormalized probability $\hat{p_i}$. The normalized probability that neuron $i$ wins the competition is given by $\tilde{p_i}$. 

Equations ~\ref{eq:mean},~\ref{eq:variance},~\ref{eq:p_unnormalized}, and~\ref{eq:p_normalized} define an analytical, differentiable approximation to the probability of a neuron winning the competition in a WTA given the input layer binary activity vector ${\bf z^{in}}$. Since the WTAs in a layer are conditionally independent given the input layer activity, these equations allow us to obtain a differentiable expression for the winning probability of all the neurons in a layer with multiple WTAs. In a layer with multiple WTAs, and where $\tilde{p_i}$ is the probability that neuron $i$ wins the competition in its WTA given the input layer activity ${\bf z^{in}}$, the log probability of observing a particular spike pattern ${\bf z^{out}}$ (which is a 0/1 binary vector) is
\begin{equation}
  \label{eq:analytical_approx}
  \log\left(p({\bf z^{out}} \mid {\bf z^{in}})\right) = \sum\limits_i log(\tilde{p_i}) z^{out}_i.
\end{equation}
${\bf z^{out}}$ has to be a valid spike pattern, i.e, with exactly one neuron in each WTA emitting a spike. Our approximation of a layer's conditional distribution can be used to obtain an explicit expression for the distribution in Eq.~\ref{eq:distro}.

\subsection{Reparameterizing the latent variables distribution}
\label{sec:gumbel}

\begin{figure}[t]
  \centering
    \includegraphics[width=0.7\textwidth]{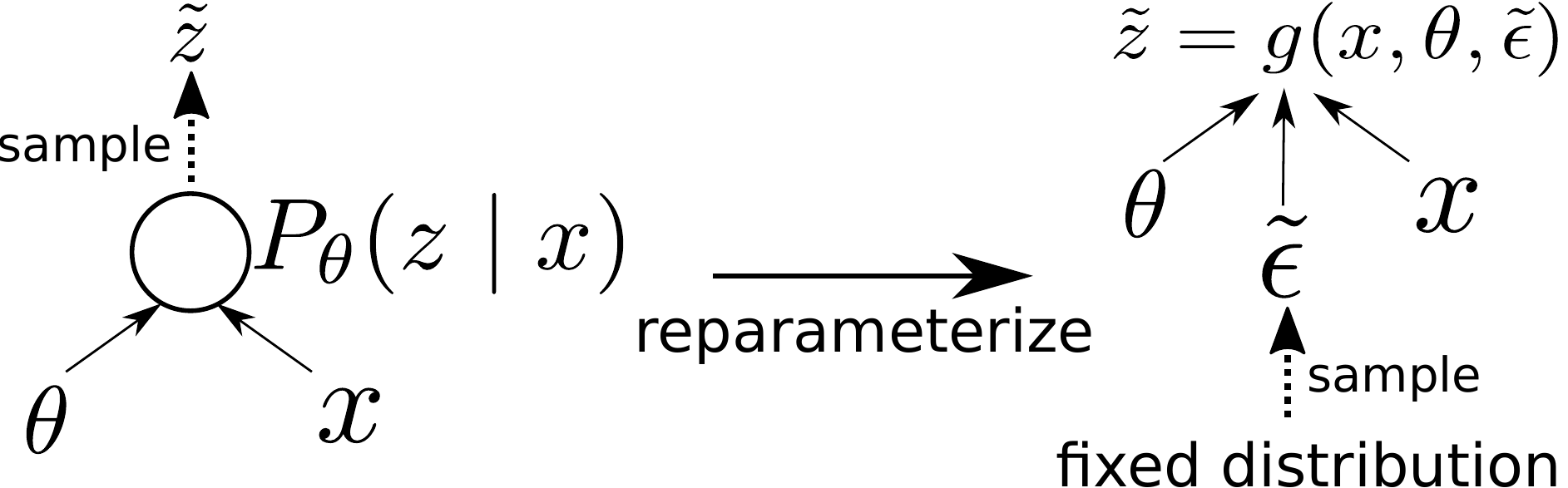}
    \caption{The reparameterization of probability distributions to obtain a differentiable relation between the samples and the distribution parameters. Instead of sampling directly from a distribution (left), the distribution is reparameterzied so that a sample can be obtained using a differentiable deterministic transformation, $g$, where randomness is injected using stochastic samples $\tilde{\epsilon}$ from a fixed distribution (right). This results in a differentiable pathway from the stochastic sample $\tilde{z}$ to the distribution parameters $\theta$ and the conditioned variable $x$. For example, instead of sampling directly from a Gaussian ${\tilde z} \sim \mathcal{N}(z;\mu(\theta,x),\sigma(\theta,x))$, we sample first from a standard normal distribution $\tilde{\epsilon}\sim \mathcal{N}(\epsilon,0,1)$ then obtain $\tilde{z}$ using $\tilde{z} = g(x,\theta,\tilde{\epsilon}) = \mu(\theta,x) + \tilde{\epsilon} \sigma(\theta,x))$ }
    \label{fig:reparam_trick}

\end{figure}

When training stochastic WTA networks containing latent variables ${\bf z}$, we typically need to evaluate expectations of the form ${\mathbb E}_{P_{\theta}({\bf z}\mid {\bf x_0})}(f({\bf z},{\bf x_0}))$, where $P_{\theta}({\bf z} \mid {\bf x_0})$ is the probability distribution over the latent variables given an observed data point ${\bf x_0}$; $\theta$ is the vector of network parameters (synaptic weights and transmission failure probabilities) which implicitly define the distribution $P$; and $f$ is a differentiable function of ${\bf z}$ and ${\bf x_0}$. $f$ can be for example the log probability of observing a particular configuration of latent variables, or it can be a variational lower bound on the data log-likelihood. Exact evaluation of this expectation is typically intractable so it is often approximated using $N$ samples from  $P_{\theta}({\bf z}\mid {\bf x_0})$:
\begin{equation}
\label{eq:sample_approx}
 {\mathbb E}_{P_{\theta}({\bf z}\mid {\bf x_0})}(f({\bf z},{\bf x_0})) \approx \frac{1}{N} \sum\limits_{{\bf z^i}\sim P_{\theta}({\bf z} \mid {\bf x_0})} f({\bf z^i},{\bf x_0}).
\end{equation}
In order to use gradient descent to maximize or minimize this expectation with respect to the the parameters $\theta$, we need to be able to estimate $\nabla_{\theta}{\mathbb E}_{P_{\theta}({\bf z}\mid {\bf x_0})}(f({\bf z},{\bf x_0}))$. Evaluating these derivatives is also analytically intractable so we have to resort to sampling approaches. One of the classical algorithms for estimating gradients of stochastic expectations with respect to the parameters of the expectation probability density is the REINFORCE algorithm~\citep{Williams92}. Recently, however, better estimators with greatly reduced variance have been introduced~\citep{Kingma_Welling13,Rezende_etal14,Ruiz_etal16}. These estimators make use of the so-called `reparameterization trick' which is illustrated in Fig.~\ref{fig:reparam_trick}. Through reparameterization, samples from a probability distribution can be written as a differentiable transformation of the distribution parameters and stochastic samples from a fixed standard distribution. Thus, one can directly use the chain rule to obtain the derivative of the sample-based estimate of the expectation in Eq.~\ref{eq:sample_approx} with respect to the distribution parameters $\theta$ since $f$ is differentiable, and through the reparameterization trick, we have a differentiable relation between the stochastic samples ${\bf z^i}$ and the distribution parameters $\theta$.

Not all distributions admit such a reparameterization. Most notably, reparameterizations of discrete distributions can not be differentiable. An approximate differentiable reparameterization that works well in practice is the Gumbel-softmax reparameterization~\citep{Jang_etal17,Maddison_etal16} which we reproduce below. Given a discrete distribution described by the probability vector ${\bf p} = [p_1,\ldots,p_K]$ defining the probability of each of the distribution's $K$ outcomes, a sample can be drawn from this discrete distribution using the following non-differentiable reparameterization:
\begin{equation}
\label{eq:gumbel}
{\bf \tilde{z}} = \onehot(\argmax\limits_j(g_j + log (p_j))).
\end{equation}
${\bf \tilde{z}}$ is a one-hot vector with exactly one-entry equal to one and the others zero. The index of this entry is the sample outcome and is given by the $\argmax$ operator. $g_1,\ldots,g_K$ are independent samples from the $Gumbel(0,1)$ distribution~\citep{Gumbel_Lieblein54}. Samples from $Gumbel(0,1)$ can be obtained by first sampling from the uniform distribution $u_j\sim Uniform(0,1)$ and then transforming the uniform samples using $g_j = -log(-log(u_j))$. ${\bf \tilde{z}}$ is a one-hot $K-dim$ vector that can take one of $K$ possible values. In order to obtain a continuous differentiable reparameterization, the $\argmax$ operator is relaxed to the differentiable $softmax$ function:
\begin{equation}
\label{eq:gumbel_softmax}
\hat{z}_j = \frac{exp((g_j + log(p_j))/\tau)}{\sum_{l=1}^K exp((g_l + log(p_l))/\tau)} \quad j=1,\ldots,K.
\end{equation}
The sample vector ${\bf \hat{z}} = [\hat{z}_1,\ldots,\hat{z}_K]$ is now differentiable with respect to the parameters of the discrete distribution, ${\bf p}$. ${\bf \hat{z}}$ is a sample from a continuous distribution that approaches the original discrete distribution as the temperature parameter $\tau$ approaches $0$. While training the network, $\tau$ is gradually annealed towards $0$. Instead of sampling ${\bf z^i}$ directly from $P_{\theta}({\bf z}\mid {\bf x_0})$, we use the probabilities of the different discrete  states of each WTA (given by Eqs.~\ref{eq:mean},~\ref{eq:variance},~\ref{eq:p_unnormalized}, and~\ref{eq:p_normalized}) in Eq.~\ref{eq:gumbel_softmax} to sample the WTA states, and use the resulting reparameterized continuous-valued samples, ${\bf \hat{z}^i}$, to evaluate the expectation in Eq.~\ref{eq:sample_approx}. When sampling from multi-layer networks such as the network in Fig.~\ref{fig:model}, we use the reparameterized continuous-valued samples of one layer to evaluate the winning probabilities of the WTAs in the next layer, and then use these winning probabilities to obtain reparameterized continuous-valued samples from the next layer. We do this layer by layer until we obtain reparameterized continuous-valued samples of all the WTAs in the network. Any differentiable cost function involving these continuous-valued samples can be optimized using backpropagation. During the annealing of $\tau$, we always keep it well above zero to avoid instabilities in training since the derivatives of the stochastic samples with respect to the network parameters diverge as $\tau$ approaches zero. In the rest of this paper, all loss functions are optimized using ADAM~\citep{Kingma_Ba14} within the Theano framework~\citep{Bastien_etal12,Bergstra_etal10}. 

In the exact case, only one neuron can win the competition in a WTA. Samples obtained using the continuous relaxation in Eq.~\ref{eq:gumbel_softmax}, however, yield a continuous-valued activity vector for the WTA circuit, where neurons that have a higher probability of winning are more likely to have higher activities in the sample vectors. This can be understood as approximating the hard WTA circuit by a soft WTA circuit~\citep{Douglas_Martin04} where the winning neuron does not completely shut down the activity in the other WTA neurons. We only use this soft WTA mechanism during training to allow gradients to flow back through stochastic samples. During testing, we use the hard WTA mechanism where only the neuron receiving the largest input emits a spike, and use the exact network dynamics with stochastic synaptic transmission failures to generate samples from the network instead of the approximation in Eq.~\ref{eq:p_normalized}. 

\subsection{Networks with realistic neuronal and synaptic dynamics}
\label{sec:lif_description}
The model presented in the previous subsections abstracts away the dynamical nature of a realistic spiking network in favor of an analytically tractable behavior with no real temporal dynamics. Analyzing the behavior of a multi-layer spiking network with realistic temporal dynamics is particularly challenging as an analytical formulation of the network behavior is typically not available. It has been common practice to first develop an abstract analytically tractable model then map the parameters of the abstract model to that of a realistic spiking network model~\cite{Buesing_etal11,Pecevski_etal11,Nessler_etal13}. The discrepancy between the behavior of the two models is then empirically investigated to validate the ability of the abstract model to capture the relevant behavior of the more realistic model. This approach extends to training spiking network with realistic temporal dynamics where it has been very effective in training feedforward networks for classification tasks~\cite{OConnor_etal13,Diehl_etal15,Cao_etal15,Hunsberger_Eliasmith15}. This approach first trains an abstract artificial neural network using standard backpropagation techniques, then the weights are mapped to the realistic spiking network. 

In this subsection, we introduce a biologically realistic spiking network model and introduce a modification to the training procedure of the abstract model in order to obtain similar behavior from the two models once the parameters are mapped from the abstract model to the biologically realistic spiking model. We use LIF neurons with conductance-based synapses. One LIF neuron is used for each neuron in the abstract model. The inter-layer connection patterns remain the same. Within the same layer, LIF neurons in the same WTA circuit have all-to-all inhibitory connections to obtain competitive behavior. The full neuron model as well as the parameters used in the subsequent simulations are given in Appendix~\ref{app:spiking_model}.

There are two ways the behavior of the network model using LIF neurons can deviate from the behavior of the abstract model:
\begin{enumerate}
\item None of the LIF neurons in a WTA spike. LIF neurons have a spiking threshold mechanism and no LIF neuron in a WTA will spike if all of them receive sub-threshold input. In the abstract model, the neuron with the largest input in the WTA will spike, regardless of the magnitude of that input.
\item The winning LIF neuron in a WTA is decided before all the WTAs in the previous layer have spiked. The asynchronous dynamics of the LIF spiking network can allow a WTA to decide a winner based only on partial input from the previous layer. This decision might be different from the winner decision obtained using the synchronous dynamics that evaluate the layers in a strict sequence where the neurons in one layer take into account input from all the winners in the previous layer.
\end{enumerate}
We ameliorate the effects of the first source of deviation by adding an additional cost term to the training cost function that is minimized when the winning neuron in each WTA has an above-threshold mean input. More precisely, this additional cost term for an individual WTA has the form
\begin{equation}
  \label{eq:TCC}
  threshold\_crossing\_cost = \begin{cases}
    -\max\limits_i \mu(t_i) & \text{if \quad $\max\limits_i \mu(t_i)  < H$} \\
  0       & \text{otherwise}
\end{cases}
\end{equation}
where $\mu(t_i)$ is the mean input to the $i^{th}$ neuron in the WTA (Eq.~\ref{eq:mean}) and $H$ is chosen to to be slightly larger than the total input needed to trigger a spike in a LIF neuron. Minimizing this cost term encourages the winning neuron in each WTA to have a super-threshold input allowing us to map the weights of the abstract model directly to a network of LIF neurons. The first source of deviation between the abstract model and the LIF network is not completely eliminated, though, as the additional cost term only affects the mean input. There is thus still a possibility that for some synaptic failure patterns, all neurons in a WTA will get sub-threshold input and all fail to spike.

The second source of deviation can be ameliorated by scaling the synaptic weights by a factor greater than unity after training. Scaling the weights does not change the winning probabilities in each WTA. Since the winning neurons are typically receiving a positive, super-threshold input due to the cost term in Eq.~\ref{eq:TCC}, this scaling results in all the winning neurons receiving a strong positive input that makes them spike in close temporal proximity. The network operation thus becomes similar to a synfire chain~\cite{Abeles82} where a synchronous volley of spikes from the input layer triggers an almost synchronous volley of spikes from the next layer and so on~\cite{Rotter_Aertsen98}. Simulations of the LIF spiking network were carried out using NEST~\cite{Gewaltig_Diesman07}.

\section{Results}
\subsection{Structured output prediction}
\begin{figure}[t]
  \centering
  \begin{subfigure}[b]{0.3\textwidth}
    \includegraphics[width=\textwidth]{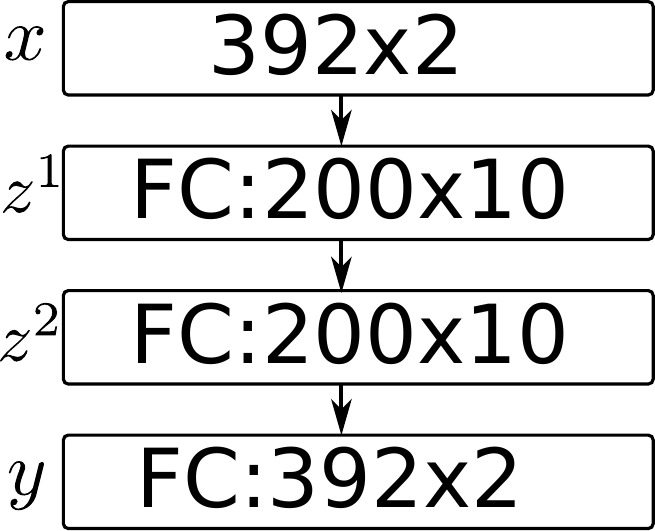}
    \subcaption{}
    \label{fig:struct_predict}
  \end{subfigure}
  \quad
  \begin{subfigure}[b]{0.3\textwidth}
    \includegraphics[width=\textwidth]{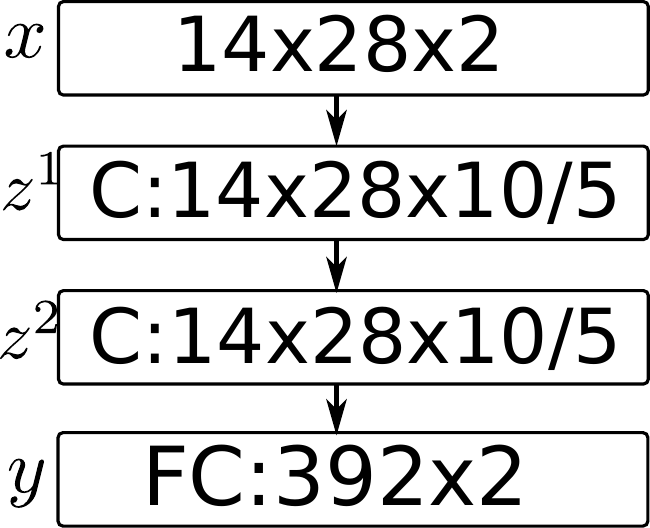}
    \subcaption{}
    \label{fig:struct_predict_conv}
  \end{subfigure}
\\

  \begin{subfigure}[b]{0.34\textwidth}
    \includegraphics[width=\textwidth]{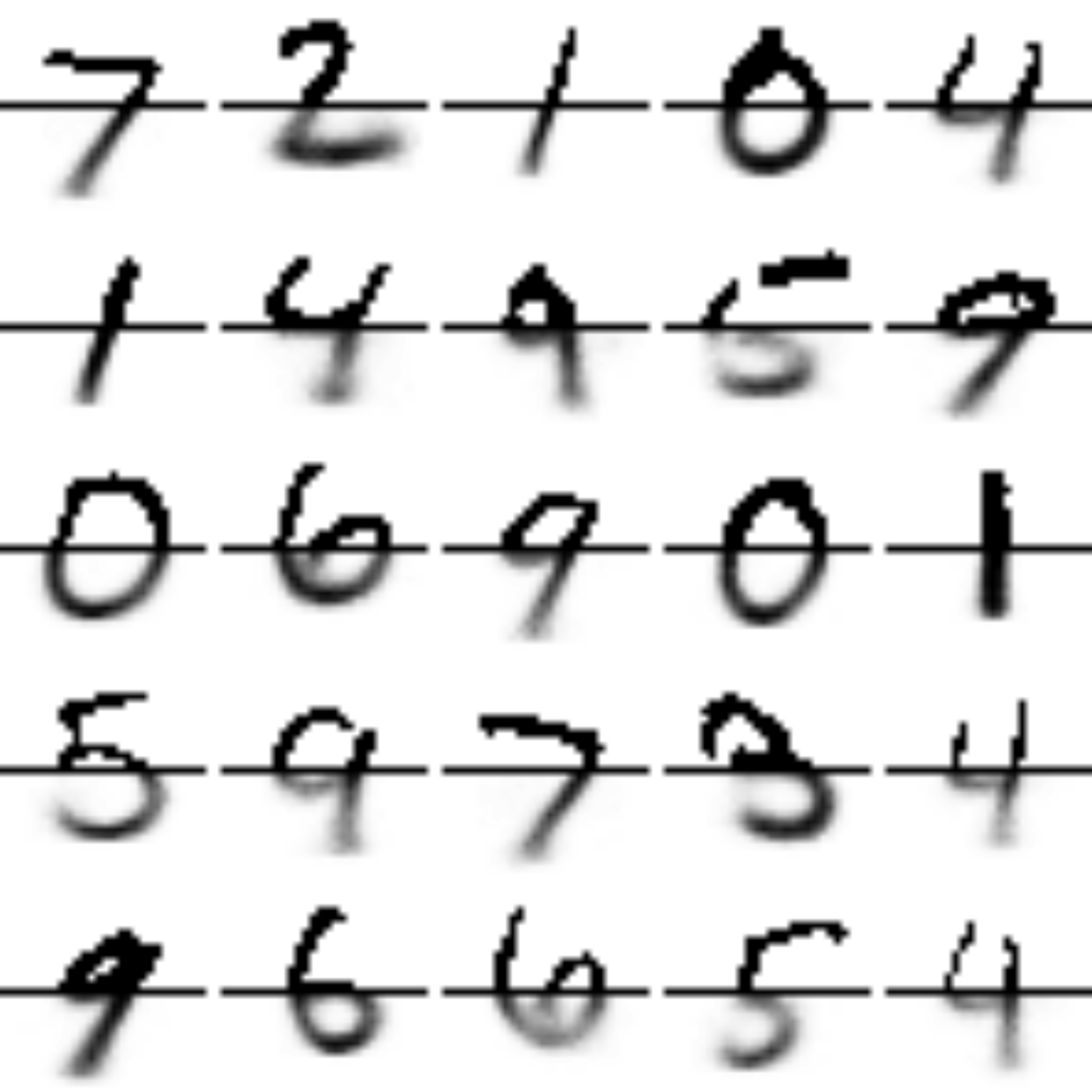}
    \subcaption{}
    \label{fig:struct_predict_results_abstract}
   \end{subfigure}
  \quad
  \begin{subfigure}[b]{0.34\textwidth}
    \includegraphics[width=\textwidth]{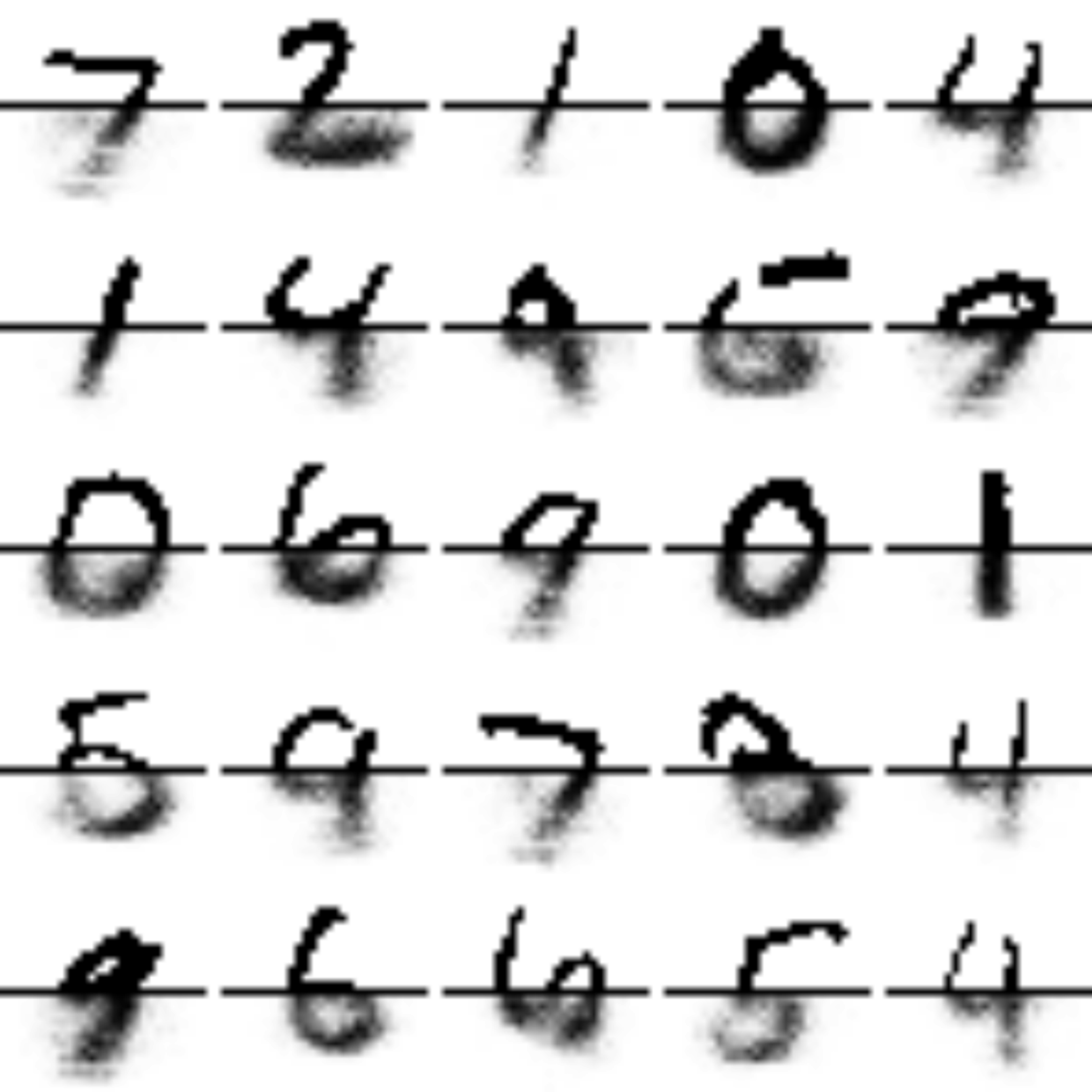}
    \subcaption{}
    \label{fig:struct_predict_results_lif}
   \end{subfigure}

  \caption{(\subref{fig:struct_predict},\subref{fig:struct_predict_conv}) Networks used in structured output prediction task. $C:A\times B \times C/D$ indicates a convolutional layer with $C$ feature maps, $A \times B$ spatial dimensions, and $D\times D$ kernel size. $FC:A \times B$ indicates a fully connected layer with $A$ WTAs and $B$ neurons in each WTA. (\subref{fig:struct_predict_results_abstract},\subref{fig:struct_predict_results_lif}) Network prediction of the lower half of sample test digits. Each digit prediction used 100 samples. Horizontal line in each digit separates the upper (input) from the bottom (predicted) half. (\subref{fig:struct_predict_results_abstract}) Samples generated using the abstract fully connected network in \subref{fig:struct_predict}. (\subref{fig:struct_predict_results_lif}) Samples generated using a LIF network with the same structure as \subref{fig:struct_predict}. The abstract network in \subref{fig:struct_predict} was trained with the threshold-crossing constraint and the parameters mapped to the LIF network.}
\end{figure}

We first validate the soundness of our approximation of the WTA winning probabilities in a structured output prediction task where the goal is to predict the lower half of an MNIST image given the upper half. The MNIST dataset contains 70000 $28\times28$ grayscale images of handwritten digits split into three groups of 50000, 10000, and 10000 images for training, validation, and testing respectively. As in ref.~\citep{Raiko_etal14,Gu_etal14}, we use a binarized version of the MNIST images obtained by thresholding the pixel intensities. We use two different types of feedforward networks illustrated in Figs.~\ref{fig:struct_predict} and~\ref{fig:struct_predict_conv} to predict the lower image half. The networks differ in the structure of the two latent variable layers: $z^1$ and $z^2$. $z^1$ and $z^2$ can either be fully-connected layers that receive all-to-all connections from the previous layer (Fig.~\ref{fig:struct_predict}), or convolutional layers where each neuron has a local receptive field (Fig.~\ref{fig:struct_predict_conv}). The final prediction layer, $y$, is always fully connected to the previous layer. We represent the binary inputs and outputs using WTA circuits with 2 neurons where the spike from one neuron codes for binary 0 and the spike from the other neuron for binary 1. Given a training pair $x_{lower}$ and $x_{upper}$, the goal is to maximize the log probability of predicting the lower image half given the upper half, $log(P_{\theta}(y=x_{lower} \mid x=x_{upper}))$, across all training pairs where $P_{\theta}$ is the probability distribution encoded by the network in Fig.~\ref{fig:struct_predict} . Exact evaluation of this training quantity would require marginalizing over the hidden variables $z_1$ and $z_2$. Instead, we estimate it using $N$ samples from the hidden variables per example:
\begin{equation}
\frac{1}{N} \sum\limits_{z_1^i \sim P_{\theta}(z_1 \mid x_{lower}),z_2^i \sim P_{\theta}(z_2 \mid z_1^i)} log(P_{\theta}(y=x_{lower} \mid z_2^i)),
\end{equation}
where we use $N=1$ sample during training. We generate samples using the continuous reparameterization from Eq.~\ref{eq:gumbel_softmax}. This way we can backpropagate errors through these samples to the network parameters and to samples from earlier layers. During training, we use an annealing schedule to lower the temperature of the softmax used to approximate the hard WTA mechanism. The learning rate also decays during training. The validation set was used to tune the temperature and learning rate schedules. Even though synaptic transmission failure can be a trainable parameter, we kept it fixed at $0.5$ for all synapses and only train the synaptic weights.

The network is tested using samples generated in two different ways: samples generated using the approximate analytical distribution used during training where temperature is set to zero, and samples generated using the exact network dynamics, i.e, using the hard WTA mechanism and synaptic transmission failures. In both cases, 100 samples were used to evaluate the log likelihood of the lower half of the image given the upper half for each test digit. The results are shown in the first two rows of Table~\ref{tab:structured_prediction}. The networks performs better on the test set using the approximate analytical distribution used during training. The difference in performance compared to the exactly generated samples is slight, however, indicating that the approximation distribution reasonably matches the exact distribution over the test set, and that effective learning can be achieved using the approximate distribution. The convolutional network significantly lags the fully connected network in performance. This is expected as the local structure at a particular point in the upper digit half has very little to do with the local structure at the corresponding point in the lower half. The test set negative log likelihood of $61.8\pm 0.074\, nats$ achieved by the fully-connected network version is slightly worse than the test set negative log likelihood of 58.5 nats achieved with previous stochastic networks~\citep{Jang_etal17} that use discrete neural variables with activation noise rather than synaptic transmission noise.

We retrained the networks in Figs.~\ref{fig:struct_predict} and~\ref{fig:struct_predict_conv} using the threshold-crossing constraint in Eq.~\ref{eq:TCC}. As shown in Table~\ref{tab:structured_prediction}, this leads to a degradation of performance as the network parameters now have an additional constraint that is unrelated to the objective. We mapped the abstract WTA networks trained using the threshold-crossing constraint to a spiking network with LIF neurons and evaluated the log-likelihood of the lower image half using 50 samples. These samples were generated by simulating the network of LIF neurons 50 times using different synaptic failure patterns. There is a significant degradation in performance when going to an asynchronous network of LIF neurons due to the two sources of deviation outlined in subsection~\ref{sec:lif_description}. The performance degradation is more severe in the case of convolutional networks due to their local receptive fields structure; by using a $5\times 5$ kernel, a neuron in a convolutional layer receives input from only 25 active neurons. It is thus more probable that a synaptic failure pattern will cause the neuron to receive a sub-threshold input and fail to spike, compared to the fully connected network where each neuron receives input from 392 or 200 active neurons. Figures~\ref{fig:struct_predict_results_abstract} and~\ref{fig:struct_predict_results_lif} shows some examples of the network prediction of the lower halves of MNIST test set digits.

\begin{table}[h]
  \caption{Performance of structured output prediction networks (negative log-likelihood). Mean and standard deviation from 10 runs. Only the result of one run is reported for the LIF networks.  }
  \label{tab:structured_prediction}
  \centering
  \begin{tabular}[t]{m{64mm} m{39mm}m{38mm}}
    \toprule
         & Samples generated using approximate analytical distribution   &  Samples generated using exact network dynamics \\
    \midrule
    {\bf Fully-connected} abstract network (Fig.~\ref{fig:struct_predict}) & $60.1 \pm 0.067\,nats$ & $61.8\pm 0.074\, nats$ \\
    {\bf Fully-connected} abstract network with threshold-crossing constraint & $ 62.86\pm 0.29\,nats$ & $65.68\pm 0.27\,nats$ \\
    {\bf Convolutional} abstract network (Fig.~\ref{fig:struct_predict_conv}) & $66.98 \pm 0.10\,nats$ & $71.77\pm 0.44\, nats$ \\
    {\bf Convolutional} abstract network with threshold-crossing constraint & $70.09 \pm 0.35\,nats$ & $73.6\pm 0.39\, nats$ \\
    {\bf Fully-connected} LIF network & - & $89.8\, nats$ \\
    {\bf Convolutional} LIF network & - & $118.5\, nats$ \\    
    \bottomrule
  \end{tabular}
\end{table}

\subsection{Learning generative models using variational autoencoders}
\begin{figure}[t]
  \centering
  \begin{subfigure}[b]{0.7\textwidth}
    \includegraphics[width=\textwidth]{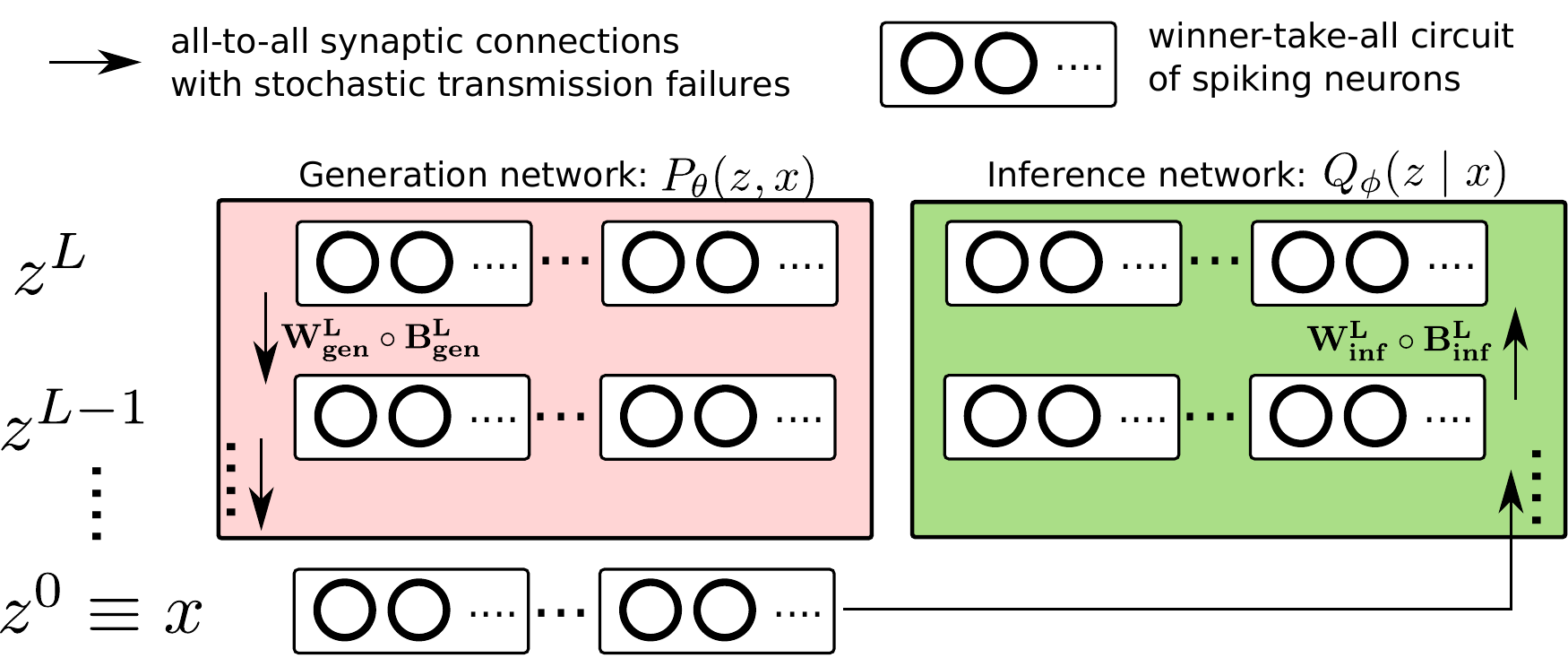}
    \subcaption{}
    \label{fig:model_vae}
  \end{subfigure}
  \\
   \begin{subfigure}[b]{0.35\textwidth}
    \includegraphics[width=\textwidth]{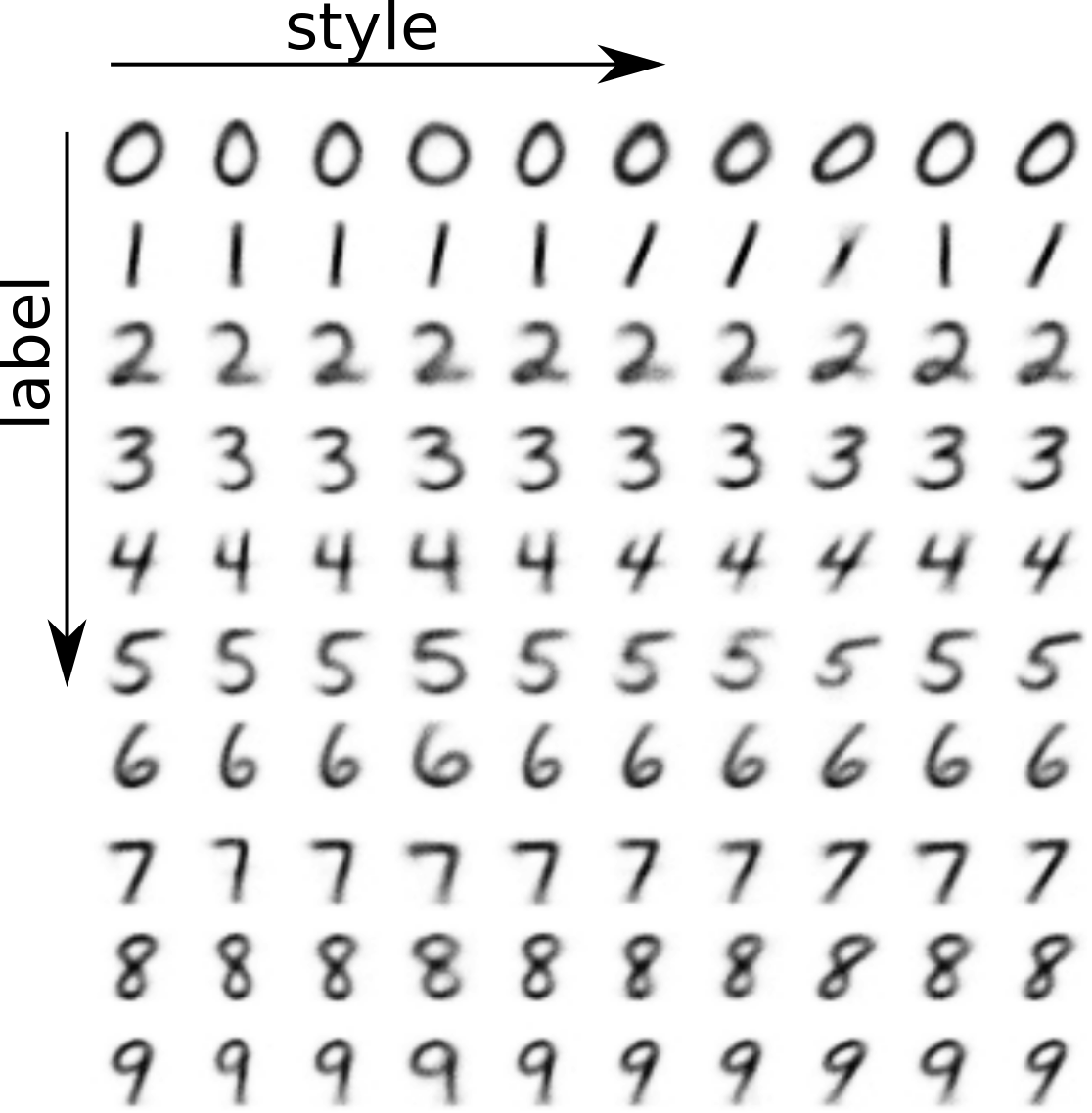}
    \subcaption{}
    \label{fig:style_label}
   \end{subfigure}
   \quad
   \begin{subfigure}[b]{0.35\textwidth}
    \includegraphics[width=\textwidth]{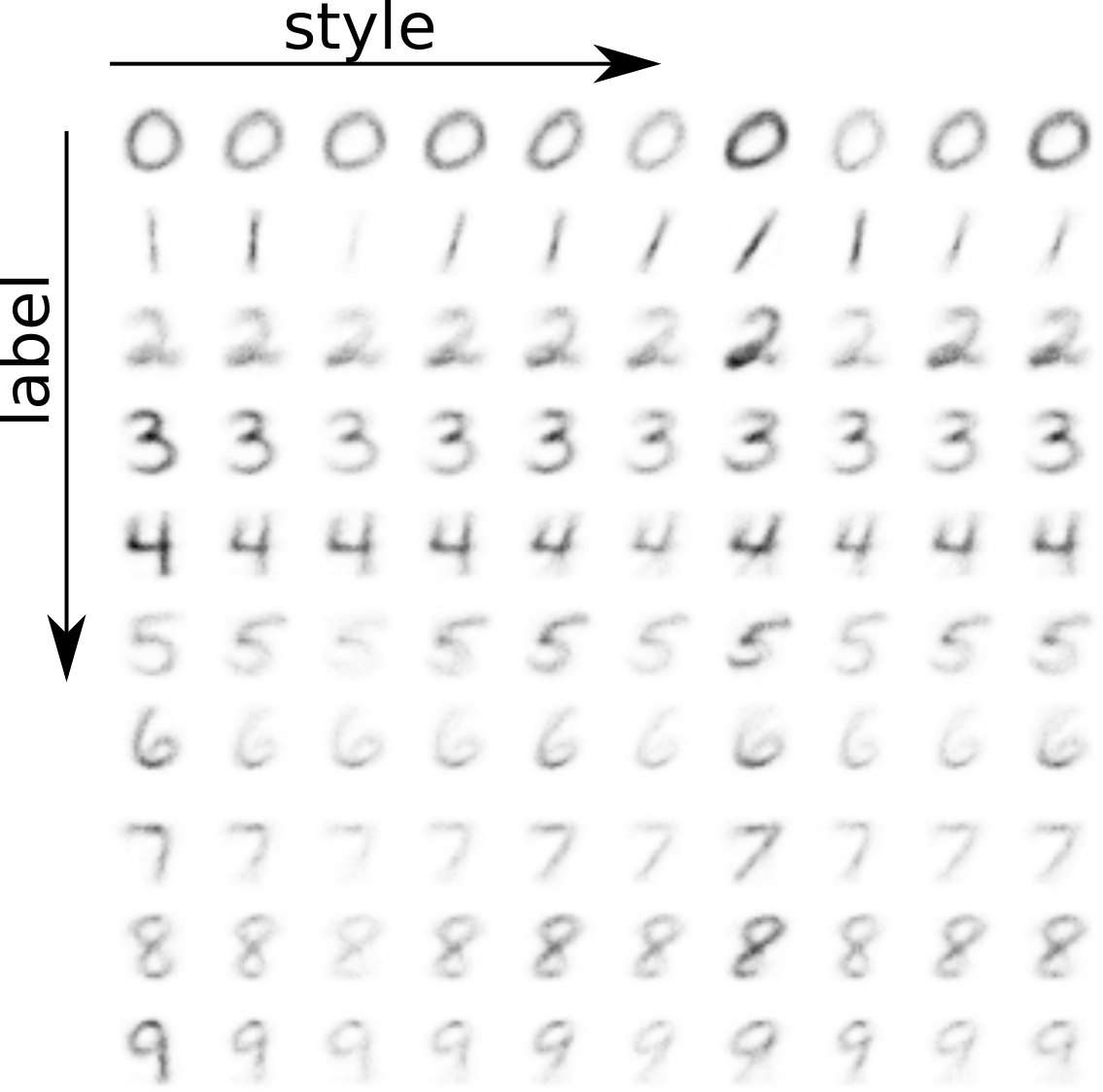}
    \subcaption{}
    \label{fig:style_label_lif}
   \end{subfigure}

  \caption{(\subref{fig:model_vae}) Structure of the variational auto-encoder network. (\subref{fig:style_label},\subref{fig:style_label_lif}). Digits generated using the generative network with separate style and label WTAs in the top layer. Each digit is constructed by fixing the top layer activity, then averaging across 300 samples. (\subref{fig:style_label}) Digits generated by the abstract network. (\subref{fig:style_label_lif}) Digits generated by a LIF network having the same structure. The corresponding abstract network was trained using the threshold-crossing constraint, then the parameters were directly mapped to the LIF network. The digits are less well-defined since sometimes, neurons in one WTA receive sub-threshold input and all fail to spike, making it more likely that the visible layer WTAs also fail to spike.}
\end{figure}

Our goal is to learn a generative model for a set of data points with discrete values ${{\bf X}=\{{\bf x^1},{\bf x^2},\dots,{\bf x^N}\}}$ where ${\bf x^i} \in \{1,2,\ldots ,k\}^D$. We use a feedforward network of WTA circuits connected by stochastic synapses to encode the generative distribution. This is the generation network shown in Fig.~\ref{fig:model_vae}. We lump all the latent variables together in one vector ${\bf z} = [{\bf z^1},\ldots,{\bf z^L}]$ and the distribution over the network spike patterns is given by $P_{\theta}({\bf x},{\bf z})$ where $\theta$ represents the generation network parameters. For the prior on the top layer variables, ${\bf z}^L$, we choose the uniform prior where each neuron has the same probability to win the competition in its WTA. Our goal is to maximize the data log likelihood $\log(P_{\theta}({\bf X}))$. We use variational methods to maximize this likelihood. For a particular input point, ${\bf x}$,
\begin{align}
\log(P_{\theta}({\bf x})) = \sum\limits_{{\bf z}}P_{\theta}({\bf z} \mid {\bf x}) \log(P_{\theta}({\bf x})) = \sum\limits_{{\bf z}}P_{\theta}({\bf z} \mid {\bf x}) \log\left(\frac{P_{\theta}({\bf x},{\bf z})}{P_{\theta}({\bf z} \mid {\bf x})}\right).
\end{align}
$P_{\theta}({\bf x},{\bf z})$ is easy to evaluate using Eqs~\ref{eq:distro} and ~\ref{eq:analytical_approx}. The posterior distribution $P_{\theta}({\bf z} \mid {\bf x})$ is, however, intractable. We approximate the intractable posterior by the parameterized distribution $Q_{\phi}({\bf z} \mid {\bf x})$. We use a multi-layer network of WTAs connected using stochastic synapses to implement $Q_{\phi}({\bf z} \mid {\bf x})$. This is the inference/recognition network shown in Fig.~\ref{fig:model_vae}. It has an analogous structure to the generation network except that the connections between layers are going in the opposite direction. The weights and synaptic transmission failure probabilities in the two networks are independent. Given $x$, the inference network induces a probability distribution $Q_{\phi}({\bf z} \mid {\bf x})$ over the latent variables which can be expressed in an analogous fashion to the generative distribution using Eqs.~\ref{eq:distro} and~\ref{eq:analytical_approx}.  The distribution parameters $\phi$ are the weights and synaptic transmission failure probabilities in the inference network. Note that unlike typical variational auto-encoder architectures, the latent variable layers are directly connected and there are no deterministic layers in the network. The data log-likelihood can be written as

\begin{align}
  \log(P_{\theta}({\bf x})) &= \sum\limits_{{\bf z}}Q_{\phi}({\bf z} \mid {\bf x}) \log\left(\frac{P_{\theta}({\bf x},{\bf z})}{P_{\theta}({\bf z} \mid {\bf x})}\frac{Q_{\phi}({\bf z}\mid{\bf x})}{Q_{\phi}({\bf z} \mid {\bf x})}\right) \\
  &= KL(Q_{\phi}({\bf z} \mid {\bf x}) \|  P_{\theta}({\bf z} \mid {\bf x})) + \sum\limits_{{\bf z}}Q_{\phi}({\bf z} \mid {\bf x}) \log\left(\frac{P_{\theta}({\bf x},{\bf z})}{Q_{\phi}({\bf z} \mid {\bf x})}\right) \\
  &= KL(Q_{\phi}({\bf z} \mid {\bf x}) \|  P_{\theta}({\bf z} \mid {\bf x})) + \mathcal{L}({\bf x};\theta,\phi).
\end{align}
$KL$ is the Kullback-Leibler divergence between two distributions and since it is non-negative, $\mathcal{L}$ is a lower bound on the data log-likelihood given by $\mathcal{L}={\mathbb E}_{Q_{\phi}({\bf z} \mid {\bf x})}\left[\log\left(\frac{P_{\theta}({\bf x},{\bf z})}{Q_{\phi}({\bf z} \mid {\bf x})}\right)\right]$. Since it is intractable to evaluate $\mathcal{L}$  exactly due to the required summation over all ${\bf z}$, we approximate it using $N$ samples from the inference network: 

\begin{equation}
\label{eq:ELBO}
\mathcal{L} \approx \frac{1}{N} \sum\limits_{{\bf z^i} \sim Q_{\phi}({\bf z}\mid{\bf x})} \log\left(\frac{P_{\theta}({\bf x},{\bf z^i})}{Q_{\phi}({\bf z^i} \mid {\bf x})}\right).
\end{equation}
We maximize this sample-based lower bound through gradient descent on the parameters $\theta$ and $\phi$, where we use the differentiable Gumbel-softmax reparameterization described in section~\ref{sec:gumbel} to backpropagate errors through the stochastic samples. A common problem with training variational auto-encoders with multiple layers of stochastic hidden variables is the collapse of the latent approximate posterior $Q_{\phi}({\bf z} \mid {\bf x})$ towards the latent prior $P_{\theta}({\bf z})$ which stops the latent variables in upper layers from encoding information about the input during training~\citep{Sonderby_etal16,Chen_etal16a}. We adopt the solution proposed in ref.~\citep{Sonderby_etal16} in which we initially only maximize the reconstruction likelihood and gradually morph the cost function to optimize the variational lower bound $\mathcal{L}$. Our optimization objective is thus 
\begin{equation}
\mathcal{J} = \frac{1}{N} \sum\limits_{{\bf z^i} \sim Q_{\phi}({\bf z}\mid{\bf x})} \log\left(P_{\theta}({\bf x} \mid {\bf z^i})\right) + \beta \log\left(\frac{P_{\theta}({\bf z^i})}{Q_{\phi}({\bf z^i} \mid {\bf x})}\right),
\end{equation}
where $\beta$ gradually increases from $0$ to $1$ during training. When $\beta=1$, $\mathcal{J}$ is equivalent to the right-hand side of Eq.~\ref{eq:ELBO}, i.e, it becomes a sample-based estimate of $\mathcal{L}$.

We apply the variational auto-encoder network based on WTA circuits and stochastic synapses to learning a generative model of MNIST digits. As in the structured output prediction task, we anneal the temperature of the softmax used in the Gumbel-softmax continuous reparameterization as well as the learning rate during training. We use $N=1$ samples per data point during training.

We trained a network with three stochastic fully-connected hidden layers of sizes $10{\times}10$, $20 {\times} 10$, and $30 {\times} 10$ going from the deepest layer downward (An $A {\times} B$ layer has $A$ WTAs with $B$ neurons in each WTA) in order to learn a generative model of the training set of MNIST digits. These layers were followed by the visible layer of $784{\times}2$ neurons. We used a uniform prior in the top layer where each neuron has an equal chance to win the competition in its WTA and used a fixed transmission failure probability of $0.5$.  After training, we evaluated the variational lower bound using the approximate distribution over WTA winners and $N=50$ samples in Eq.~\ref{eq:ELBO}. The samples were also drawn from the approximate distribution. This evaluation was done on the test set. Under the approximate distribution, we obtain a variational lower bound of $-98.9 \pm 0.3\,nats$ (mean and standard deviation from 10 runs) which is on par with discrete stochastic neural networks that use activation noise rather than synaptic transmission noise~\citep{Jang_etal17}. 

Estimating the variational lower bound using the exact network dynamics is particularly difficult in our case as exact evaluation of $P_{\theta}({\bf x},{\bf z})$ and $Q_{\phi}({\bf z} \mid {\bf x})$ is intractable. Theoretically, we can estimate these probability distributions using samples, but we observed that the number of samples needed to reliably estimate the exact $P_{\theta}({\bf x},{\bf z})$ and $Q_{\phi}({\bf z} \mid {\bf x})$ is computationally prohibitive. Note that this problem does not arise in conventional variational auto-encoders as they have exact expressions for the conditional log probability of each layer's activity. There are two steps to our approximation of the winning probability of the neurons in a WTA: the first step assumes the input to each neuron is Gaussian, the second step approximates the intractable integration in Eq.~\ref{eq:intractable} with the tractable integration in Eq.~\ref{eq:approx_integ}. The first step is practically unavoidable, as the exact distribution of the input to a neuron is an intractable discrete distribution where each point corresponds to one configuration of synaptic failures. We can dispense with the second step, however, by numerically evaluating the integration in Eq.~\ref{eq:intractable}. 

Our estimate of the variational lower bound that is more faithful to the exact network dynamics uses exactly generated samples ${\bf z^i}$ in Eq.~\ref{eq:ELBO}, i.e, samples of ${\bf z}$ generated by sampling the synaptic transmission failure probabilities, and it uses numerical integration of Eq.~\ref{eq:intractable} to obtain accurate winning probabilities in each WTA. The accurate winning probabilities are used  to evaluate $P_{\theta}({\bf x},{\bf z})$ and $Q_{\phi}({\bf z} \mid {\bf x})$. Under this estimate, the variational lower bound is $-104 \pm 1.4\, nats$ which is slightly worse than the lower bound evaluated using the approximate distribution used during training.

\begin{table}[h]
  \caption{Variational lower bound evaluated for a network with 3 hidden layers of latent variables. Mean and standard deviation from 10 runs. }
  \label{tab:vae}
  \centering
  \begin{tabular}[t]{m{64mm} m{39mm}m{38mm}}
    \toprule
         & Evaluation using approximate samples and approximate analytical distribution   &  Evaluation using exact samples and numerical integration of Eq.~\ref{eq:intractable} \\
    \midrule
    {\bf Training} without threshold-crossing constraint & $-98.9 \pm 0.3\,nats$ & $-104 \pm 1.4\, nats$ \\
    {\bf Training} with threshold-crossing constraint & $-102.46 \pm 0.18\,nats$ & $-109.12\pm 0.37\, nats$ \\
    \bottomrule
  \end{tabular}
\end{table}

We repeated the training and evaluation steps for the same network, but using the threshold crossing constraint. As in the structured output prediction task, enforcing the threshold crossing constraint reduced the network performance. The results are summarized in Table~\ref{tab:vae}. The parameters of the abstract network using the threshold crossing constraint can be directly mapped to the parameters of a LIF network. However, evaluating the variational lower bound in the LIF network is computationally intractable as it would require an infeasibly large number of samples where acquiring each sample involves the simulation of a large system of differential equations. Instead, we visually evaluate the LIF network performance on a generative task with style-label separation: we trained a variational auto-encoder network in which a subset of the WTAs in the top layer were forced to represent the MNIST image label. The network had three stochastic hidden layers of sizes $20{\times}10$, $30 {\times} 10$, and $30 {\times} 10$ going from the deepest layer downward. The top layer prior used during training was label dependent: 5 of the 20 WTAs in the top layer had a delta function prior centered on the neuron whose index corresponds to the label of the current training example, while the remaining 15 WTAs used a uniform prior. During training, different values of the top 15 WTAs with uniform prior  will represent different variants of each digit (where the digit is specified by the other 5 WTAs). Aesthetically, each configuration of the 15 WTAs will give rise to 10 digits (0 to 9) having similar style. We chose 5 WTAs to represent the label in order to have a significant label-dependent part in the top layer activity. We trained the network twice, once without the threshold crossing constrain, and once with the threshold crossing constraint. We then mapped the parameters of the network trained with the threshold crossing constraint to a LIF network. Figs.~\ref{fig:style_label} and~\ref{fig:style_label_lif} show the images generated by the abstract network trained without the threshold crossing constraint and the images generated by the LIF network. We sample from the generative network by first forcing the 5 label WTAs to represent one label and randomly choosing a winner in the remaining 15 WTAs in the top layer, i.e, we choose a label and a random style. We then let the network dynamics sample from the remaining layers.

\subsection{Semi-supervised learning using ladder networks}
Noise plays a crucial role in several auto-encoding architectures~\citep{Vincent_etal08,Bengio_etal14,Valpola15} where it limits the information flowing between network layers, thereby forcing networks to learn more compact and general representations in an unsupervised manner. In a semi-supervised setting, the representations learned in an unsupervised manner are fed to a classifier and few labeled examples are used to train the classifier (and the underlying representations as well). By simultaneously learning representations to optimize the unsupervised loss (reconstruction error) and the supervised loss (label prediction error), the network can learn to extract the label-relevant information using only few labeled examples.

\begin{figure}[t]
  \centering
    \includegraphics[width=\textwidth]{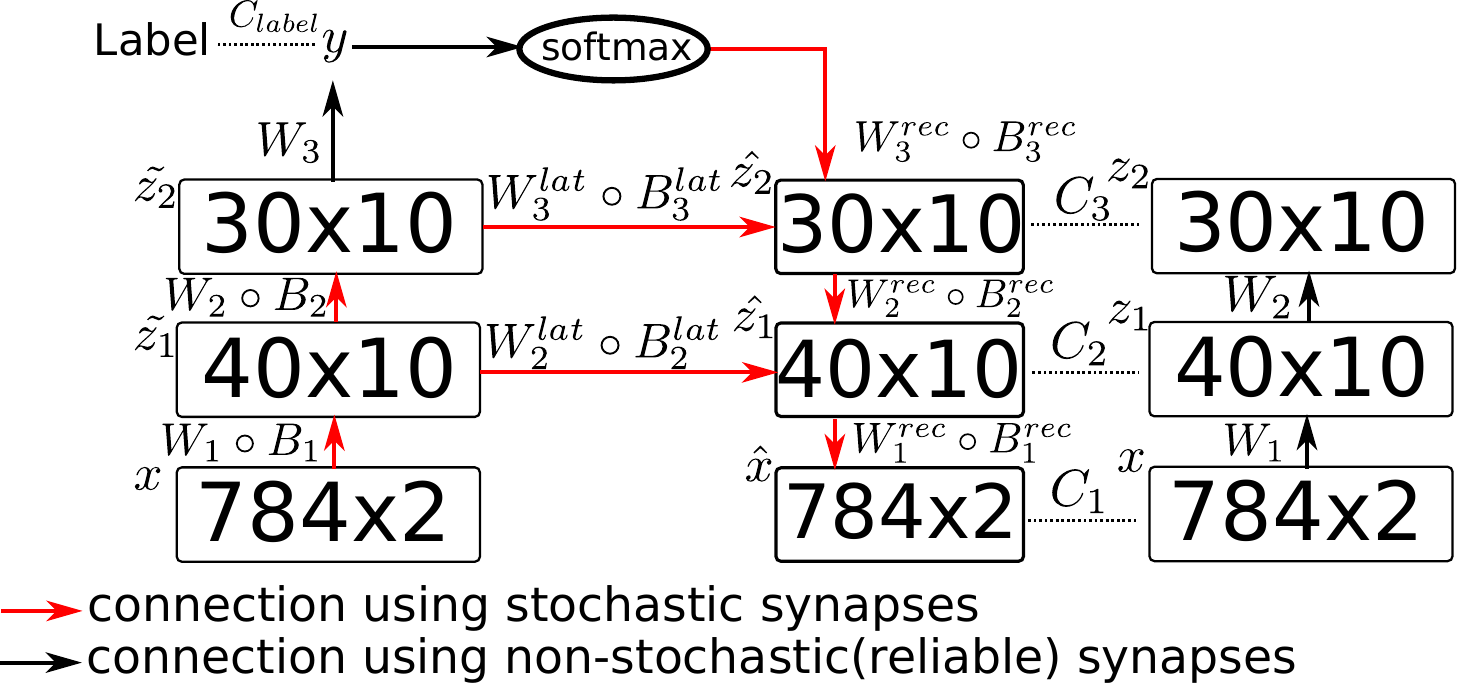}
    \caption{Ladder network used in the MNIST semi-supervised learning task. There are two input pathways, one using stochastic synapses (leftmost pathway) and one using deterministic synapses (rightmost pathway). Both pathways use the same synaptic weights. For unlabeled examples, all the weights are trained to maximize the log probability of reconstructing the activity in the deterministic input pathway using the reconstruction pathway (middle pathway). These are the cost terms $C_1$, $C_2$, and $C_3$. For labeled examples, only the weights in the input pathway, $W_1$, $W_2$, and $W_3$ are trained to maximize the probability of the correct label in the upper classifier layer with 10 units.}

    \label{fig:ladder}

\end{figure}

We used WTA networks with stochastic synapses to construct an auto-encoder and added a linear classifier on top of the deepest hidden layer. We applied this architecture to an MNIST semi-supervised learning task where only a subset of the labeled examples from the training set were used to train the classifier. The full (unlabeled) training set was used to train the auto-encoder. We used a ladder architecture~\citep{Rasmus_etal15} for the auto-encoder as shown in Fig.~\ref{fig:ladder} with lateral connections between the encoding and decoding branches. In the original ladder networks~\citep{Rasmus_etal15}, the real-valued activations in the encoding branch (leftmost branch in Fig.~\ref{fig:ladder}) are corrupted through the addition of Gaussian noise. In our case, the corruption of the encoding branch is achieved through the use of stochastic synapses. The parallel, uncorrupted branch providing the reconstruction targets (rightmost branch in Fig.~\ref{fig:ladder}) uses deterministic synapses and the same weight matrices as the corrupted encoding branch. For each training example, the encoding branch and reconstruction branch are sampled to obtain ${\bf \hat{p}_x}$, ${\bf \hat{p}_1}$, and ${\bf \hat{p}_2}$  which are the probability vectors for generating a spike in the reconstruction branch layers. ${\bf x}$, ${\bf z_1}$, and  ${\bf z_2}$ are the binary spike vectors in the corresponding layers in the clean input branch. The unsupervised reconstruction loss is then given by 
\begin{equation}
  L_{unsupervised} = - \left(<{\bf x},log({\bf \hat{p}_x})>  + <{\bf z_1},log({\bf \hat{p}_1})> + <{\bf z_2},log({\bf \hat{p}_2})>\right),
\end{equation}
where $<,>$ is the dot product operator. ${\bf y}$ is the input vector to the top classifier layer. The supervised loss at the classifier layer is the cross-entropy loss:
\begin{equation}
L_{supervised} = -log \frac{exp(y_r)}{\sum\limits_i exp(y_i)}, 
\end{equation} 
where $r$ is the input label index. During training, a fixed labeled subset of the training set is used to minimize $L_{supervised}$ while the full training set was used to minimize $L_{unsupervised}$. As in previous experiments, we anneal the temperature of the softmax used in the Gumbel-softmax continuous reparameterization as well as the learning rate during training, and keep synaptic failure probabilities fixed at $0.5$. When classifying the test set, a deterministic version of the encoder branch is used where no synaptic transmission failures occur. Table ~\ref{tab:semi-supervised} shows the classification accuracy after training with different numbers of labeled examples. 

We trained the abstract network using the threshold crossing constraint (Eq.~\ref{eq:TCC}) and then mapped the parameters to a LIF network. We then evaluated the classification performance of both networks. The classification accuracy results of the deterministic version of both network with reliable synapses are shown in table~\ref{tab:semi-supervised}. Unlike the structured output prediction and variational auto-encoder tasks, there is no drop in performance when using the threshold crossing constraint during training of the abstract network. Performance takes a hit, however, in the LIF network. In the deterministic case, the difference in activity between the LIF network and the abstract network is due to the second source of deviation outlined in section~\ref{sec:lif_description}: neurons that spike quickly before all the WTAs in the previous layer have spiked, thereby yielding a winner decision based only on partial input from the previous layer. 
\begin{table}[h]
  \caption{MNIST test set accuracy for different number of labeled training examples. For the abstract networks, means and standard deviations from 5 runs.}
  \label{tab:semi-supervised}
  \centering
  \begin{tabular}[t]{m{40mm}cccc}
    \toprule
         & 100   &  300 &  2000 & 5000    \\
    \midrule
    {\bf Abstract} network without threshold crossing constraint & $92.8 \pm 0.9\%$ & $94.7 \pm 0.26 \%$ & $95.1 \pm 0.13 \%$ &  $95.7 \pm 0.22 \%$ \\
    {\bf Abstract} network with threshold crossing constraint & $92.2 \pm 1.7\%$ & $94.7 \pm 0.34 \%$ & $95.3 \pm 0.083 \%$ &  $95.7 \pm 0.13 \%$ \\
    {\bf LIF} network & $90.0\%$ & $91.9 \%$ & $92.1 \%$ &  $93.3 \%$ \\        

    \bottomrule
  \end{tabular}
\end{table}

We are not aware of any previous work that used ladder networks with discrete neurons in a semi-supervised task. A semi-supervised learning approach that also uses discrete neurons is based on restricted Boltzmann machines and achieves a  worse performance ($92.0 \%$ accuracy with 800 labeled examples)~\citep{Larochelle_Bengio08}. A convolutional spiking network was previously trained layer-by-layer in an unsupervised manner followed by a classifier that was trained using few labelled examples~\citep{Panda_Roy16}. Our approach results in much sparser activity and outperforms this previous work when using few labelled examples. This previous work, however, reaches significantly better accuracy when the number of labelled training examples increases.

\section{Discussion}
\label{sec:conclusions}
Local neural competition mediated by inhibitory populations~\citep{Douglas_Martin92} is a ubiquitous phenomena in biological networks. This competition can take one of several forms such as recurrent excitation that causes activity in the neural population receiving the largest input to ramp up and suppress the activity in the other populations through a common inhibitory population~\citep{Douglas_Martin04}. Alternatively, competition can take the form of a race to spike among the neurons, where the neuron receiving the strongest input spikes first, triggering a volley of spikes from inhibitory neurons that suppresses activity in the local circuit. The latter mechanism, which does not keep a persistent memory of the winner, can be regulated by oscillatory inhibition where the race to spike occurs during the decaying phase of the rhythmic inhibition. This synchronizes the excitatory spikes generated in the network to periodically recurring temporal windows in which the inhibition is low~\citep{Fries_etal07}. The WTA networks described in this paper are an abstraction of the later form of competitive dynamics where the network generates a synchronous volley of spikes during each pass/cycle. 

Local neural competition can give rise to two distinct forms of the WTA mechanism. In one form, the winner's output is all-or-nothing (for example one spike), and does not encode the input to the winner; in the second form, the activity of the winning population/neuron encodes the input it received. The two forms roughly correspond to an argmax operation and a max operation, respectively. Networks employing the later form of the WTA mechanism (the max form) can be directly trained using standard backpropagation techniques. The max WTA form thus finds application in a variety of networks~\citep{Goodfellow_etal13,Srivastava13} where it effectively partitions the network into many overlapping sub-networks. Each subnetwork corresponds to a different set of winners. Such virtual partitioning improves performance as subnetworks or clusters of subnetworks can specialize to different parts of the input space~\citep{Srivastava_etal14a}. The argmax form of the WTA mechanism, however, is simpler to implement using spiking networks as the winning neuron can simply emit one spike, and information is solely encoded in the identity of the winners. Moreover, the argmax form of the WTA enables faster processing in spiking networks compared to the max form as the neuron does not have to transmit high-precision information using latency or rate codes. The virtual network partitioning property of the max form carries over to the argmax form. The argmax form is also more attractive for neuromorphic implementations as network evaluation does not involve any multiplications. Due to its many advantages, we have focused mainly on the argmax form of the WTA mechanism in this paper. The main downside of the argmax form is that it is non-differentiable. We circumvented this problem by using a softmax approximation and annealing the softmax temperature during training. Stochastic networks employing the max form of the WTA, however, can still be readily trained using the framework presented in this paper: instead of sampling the winners in each WTA, we would instead sample the input to each neuron (under the Gaussian approximation), and only allow the neuron receiving the maximum sampled input in each WTA to transmit its input while the activity of all other neurons is suppressed. The sampled input would be obtained using a reparameterization of the Gaussian distribution, allowing error information to backpropagate through the samples to the weights and transmission failure probabilities controlling the mean and variance of the Gaussian inputs to the neurons.

Our training procedure is based on backpropagation through stochastic samples. Backpropagation is biologically unrealistic for several reasons such as the need to interleave forward and backward passes, and the use of symmetric weights in the forward and backward passes. More biologically-plausible models have been proposed to address these issues that use contrastive learning in energy-based models ~\cite{Seung_etal03,Bengio_Fischer15,Scellier_Bengio17}, or that relax the symmetry requirement by using random weights in the backward pass~\cite{Lillicrap_etal16,Baldi_etal16,Nokland_etal16,Mostafa_etal17a}. These methods, however, have been applied in supervised learning settings and their performance and applicability to learning in stochastic networks is unclear. Moreover, many of these approaches are based on neurons with smooth activation functions, so their applicability to a non-rate-based spiking network such as ours is questionable. The form of a biologically-plausible learning rule for multi-layer stochastic spiking networks is thus still an open question. 

The stochastic neural network framework we developed in this paper makes use of synaptic transmission failure as the sole noise mechanism. Synaptic transmission in biological networks is highly unreliable in many cases~\citep{Allen_Stevens94,Faisal_etal08,Borst_Gerard10}. There is evidence that such unreliability is not due to biophysical constraints as synapses, even synapses with few release sites, can be highly reliable~\citep{Volgushev_etal95,Stratford_etal96}. This suggests that synaptic transmission failure could potentially serve a useful computational role~\citep{Branco_Staras09}. Stochastic network architectures typically use noisy neurons rather than noisy synapses~\citep{Ackley_etal85,Valpola15,Rezende_etal14,Kingma_Welling13}. The choice of injecting noise directly into the neurons is motivated by the analytical tractability of the neuronal noise model which often leads to simple analytical expressions for the probability distribution over possible activity patterns. In this paper, we used the synaptic noise model and derived an approximate analytical expression relating the synaptic noise to the response variability in WTA networks. Both the mean and variance of the neuron's input are now direct functions of the same synaptic weights (see Eqs.~\ref{eq:mean} and~\ref{eq:variance})) and do not utilize the separate pathways for controlling mean and variance commonly used in abstract stochastic networks~\citep{Kingma_Welling13,Rezende_etal14}. By making use of a more biophysically explicit and measurable noise mechanism such as synaptic transmission failure, the probabilistic networks presented in this paper are better-suited for developing mechanistic models of probabilistic computations in the brain compared to stochastic networks using abstract neuronal noise mechanisms.  

Using the approximate expression for the probability distribution over network states, the proposed networks are effectively trainable using recent approaches for training discrete stochastic networks~\citep{Jang_etal17,Maddison_etal16}. At test time, the difference between the expectations calculated using the network's exact and approximate probability distributions is minimal. The learning framework based on the approximate network distribution is thus accurate enough and general enough to train the proposed stochastic WTA networks in a wide range of scenarios.

The development of biologically-inspired stochastic network models that can be applied to practical problems has mainly focused on approximating Boltzmann machines~\citep{Buesing_etal11,Neftci_etal14,Neftci_etal16} using recurrently connected spiking neurons. Unbiased samples can not be quickly obtained from these models as they require running a Markov Chain Monte Carlo (MCMC) sampler. While Boltzmann distributions are quite general, multi-layer feedforward stochastic networks trained end-to-end using backpropagation have in recent years displayed superior performance as generative models~\citep{Goodfellow14,Kingma_Welling13}. Moreover, many powerful stochastic architectures for semi-supervised learning tasks~\citep{Valpola15} and auto-encoding tasks~\citep{Vincent_etal08,Bengio_etal14} do not fit into the Boltzmann machines framework. The framework of WTA networks with stochastic synapses that we present here, however, can implement a much wider class of modern stochastic network architectures, while still maintaining reasonable biological realism in the noise model and non-linearities used. 

\appendix
\section{Spiking neuron model and simulation parameters}
\label{app:spiking_model}
The dynamics of the conductance-based leaky integrate and fire neuron are described by:
\begin{align}
  C_m \dot{V}_m(t)  = &  g_L(E_l - V_m(t)) + g_{exc}(t) (E_{exc} - V_m(t)) + g_{inh}(t) (E_{inh} - V_m(t)) \\
  \dot{g}_{exc}(t)  = &  -\frac{g_{exc}(t)}{\tau_{syn}} + \sum\limits_i max(w_{i},0)\sum\limits_r\delta(t-t_i^r) \label{eq:gexc}\\
  \dot{g}_{inh}(t)  = &  -\frac{g_{inh}(t)}{\tau_{syn}} + \sum\limits_i max(-w_{i},0)\sum\limits_r\delta(t-t_i^r) \label{eq:ginh} \\
  V_m(t) \leftarrow & V_{reset} \quad\quad \text{if \quad $V_m(t) > V_{th}$}
\end{align}
where $V_m$ is the membrane potential and $g_{exc}$ and $g_{inh}$ are the excitatory and inhibitory synaptic conductances respectively. The summation over $i$ runs over all presynaptic neurons. $t_i^r$ is the time of the $r^{th}$ spike from pre-synaptic neuron $i$ and $w_i$ is the weight of the synapse from pre-synaptic neuron $i$. Positive weights trigger an increase in the excitatory synaptic conductance $g_{exc}$, while negative weights trigger an increase in the inhibitory synaptic conductance $g_{inh}$. The neuron has a refractory period of $T_{ref}$. The description of the neuron parameters and the parameter values used in simulation are given in Table~\ref{tab:parameters}. Neurons in the same WTA are connected using inhibitory connections with weight $w_{inh}^{WTA} = -200.0$. The weights of the trained abstract model are scaled by a factor of $4$ before using them in the LIF spiking network. 

\begin{table}
\centering
\begin{tabular}{lll}
  Parameter & Description & Value \\
  \hline
  $C_{m}$  & Membrane capacitance & 250.0pF \\
  $g_L$  & Leak conductance & 16.7nS \\
  $E_L$  & Leak reversal potential & -70.0mV \\
  $E_{exc}$  & Excitatory synapse reversal potential & 0.0mV \\
  $E_{exc}$  & Inhibitory synapse reversal potential & -85.0mV \\      
  $\tau_{syn}$ & Synaptic conductance time constant & 3.0 ms\\
  $V_{th}$ & Firing threshold & -65.0mV\\
  $V_{reset}$ & Reset potential & -70.0mV\\
  $T_{ref}$ & Absolute refractory period & 10.0ms\\
  \hline
\end{tabular}
\caption{Parameters of the LIF spiking neurons.}
\label{tab:parameters}
\end{table}
\FloatBarrier


\section*{Acknowledgments}
This work was supported by the Swiss national science foundation early postdoc mobility grant P2ZHP2\_164960 and by the National Science Foundation under grant 1640081, and the Nanoelectronics Research Corporation (NERC), a wholly owned subsidiary of the Semiconductor Research Corporation (SRC), through Extremely Energy Efficient Collective Electronics (EXCEL), an SRC-NRI Nanoelectronics Research Initiative.


\begin{thebibliography}{66}
\expandafter\ifx\csname natexlab\endcsname\relax\def\natexlab#1{#1}\fi

\bibitem[Abeles(1982)]{Abeles82}
Abeles, M. (1982).
\newblock Local cortical circuits. {A}n electrophysiological study.
\newblock (Springer, Berlin).

\bibitem[Ackley et~al.(1985)Ackley, Hinton, \& Sejnowski]{Ackley_etal85}
Ackley, D., Hinton, G., \& Sejnowski, T. (1985).
\newblock A learning algorithm for {Boltzmann} machines.
\newblock Cognitive science, 9, 147--169.

\bibitem[Al-Shedivat et~al.(2014)Al-Shedivat, Neftci, \&
  Cauwenberghs]{Al-Shedivat_etal16}
Al-Shedivat, M., Neftci, E., \& Cauwenberghs, G. (2014).
\newblock Learning non-deterministic representations with energy-based
  ensembles.
\newblock arXiv preprint arXiv:1412.7272.

\bibitem[Allen \& Stevens(1994)]{Allen_Stevens94}
Allen, C. \& Stevens, C. (1994).
\newblock An evaluation of causes for unreliability of synaptic transmission.
\newblock Proceedings of the National Academy of Sciences, 91, 10380--10383.

\bibitem[Baldi et~al.(2016)Baldi, Sadowski, \& Lu]{Baldi_etal16}
Baldi, P., Sadowski, P., \& Lu, Z. (2016).
\newblock Learning in the machine: Random backpropagation and the learning
  channel.
\newblock arXiv preprint arXiv:1612.02734.

\bibitem[Bastien et~al.(2012)Bastien, Lamblin, Pascanu, Bergstra, Goodfellow,
  Bergeron, Bouchard, Warde-Farley, \& Bengio]{Bastien_etal12}
Bastien, F., Lamblin, P., Pascanu, R., Bergstra, J., Goodfellow, I., Bergeron,
  A., Bouchard, N., Warde-Farley, D., \& Bengio, Y. (2012).
\newblock Theano: new features and speed improvements.
\newblock arXiv preprint arXiv:1211.5590.

\bibitem[Bengio \& Fischer(2015)]{Bengio_Fischer15}
Bengio, Y. \& Fischer, A. (2015).
\newblock Early inference in energy-based models approximates back-propagation.
\newblock arXiv preprint arXiv:1510.02777.

\bibitem[Bengio et~al.(2014)Bengio, Laufer, Alain, \& Yosinski]{Bengio_etal14}
Bengio, Y., Laufer, E., Alain, G., \& Yosinski, J. (2014).
\newblock Deep generative stochastic networks trainable by backprop.
\newblock In International Conference on Machine Learning, pp. 226--234.

\bibitem[Benjamin et~al.(2014)Benjamin, Gao, McQuinn, Choudhary,
  Chandrasekaran, Bussat, Alvarez-Icaza, Arthur, Merolla, \&
  Boahen]{Benjamin_etal14}
Benjamin, B.~V., Gao, P., McQuinn, E., Choudhary, S., Chandrasekaran, A.~R.,
  Bussat, J., Alvarez-Icaza, R., Arthur, J., Merolla, P., \& Boahen, K. (2014).
\newblock Neurogrid: A mixed-analog-digital multichip system for large-scale
  neural simulations.
\newblock Proceedings of the {IEEE}, 102, 699--716.

\bibitem[Bergstra et~al.(2010)Bergstra, Breuleux, Bastien, Lamblin, Pascanu,
  Desjardins, Turian, Warde-Farley, \& Bengio]{Bergstra_etal10}
Bergstra, J., Breuleux, O., Bastien, F., Lamblin, P., Pascanu, R., Desjardins,
  G., Turian, J., Warde-Farley, D., \& Bengio, Y. (2010).
\newblock Theano: a {CPU} and {GPU} math expression compiler.
\newblock In Proceedings of the Python for scientific computing conference
  (SciPy), vol.~4, p.~3. Austin, TX.

\bibitem[Borst \& Gerard(2010)]{Borst_Gerard10}
Borst, J. \& Gerard, G. (2010).
\newblock The low synaptic release probability in vivo.
\newblock Trends in neurosciences, 33, 259--266.

\bibitem[Branco \& Staras(2009)]{Branco_Staras09}
Branco, T. \& Staras, K. (2009).
\newblock The probability of neurotransmitter release: variability and feedback
  control at single synapses.
\newblock Nature Reviews Neuroscience, 10, 373--383.

\bibitem[Buesing et~al.(2011)Buesing, Bill, Nessler, \& Maass]{Buesing_etal11}
Buesing, L., Bill, J., Nessler, B., \& Maass, W. (2011).
\newblock Neural dynamics as sampling: A model for stochastic computation in
  recurrent networks of spiking neurons.
\newblock PLoS computational biology, 7, e1002211.

\bibitem[Cao et~al.(2015)Cao, Chen, \& Khosla]{Cao_etal15}
Cao, Y., Chen, Y., \& Khosla, D. (2015).
\newblock Spiking deep convolutional neural networks for energy-efficient
  object recognition.
\newblock International Journal of Computer Vision, 113, 54--66.

\bibitem[Chen et~al.(2016)Chen, Kingma, Salimans, Duan, Dhariwal, Schulman,
  Sutskever, \& Abbeel]{Chen_etal16a}
Chen, X., Kingma, D., Salimans, T., Duan, Y., Dhariwal, P., Schulman, J.,
  Sutskever, I., \& Abbeel, P. (2016).
\newblock Variational lossy autoencoder.
\newblock arXiv preprint arXiv:1611.02731.

\bibitem[Dayan et~al.(1995)Dayan, Hinton, Neal, , \& Zemel]{Dayan_etal95}
Dayan, P., Hinton, G., Neal, R., , \& Zemel, R. (1995).
\newblock The {Helmholtz} machine.
\newblock Neural Computation, 7, 889--904.

\bibitem[Deneve(2008)]{Deneve08}
Deneve, S. (2008).
\newblock Bayesian spiking neurons {I}: Inference.
\newblock Neural Computation, 20, 91--117.

\bibitem[Diehl et~al.(2015)Diehl, Neil, Binas, Cook, Liu, \&
  Pfeiffer]{Diehl_etal15}
Diehl, P.~U., Neil, D., Binas, J., Cook, M., Liu, S.-C., \& Pfeiffer, M.
  (2015).
\newblock Fast-classifying, high-accuracy spiking deep networks through weight
  and threshold balancing.
\newblock In International Joint Conference on Neural Networks {(IJCNN)}.

\bibitem[Douglas \& Martin(1992)]{Douglas_Martin92}
Douglas, R. \& Martin, K. (1992).
\newblock A functional microcircuit for cat visual cortex.
\newblock Jour. Physiol., 440, 735--769.

\bibitem[Douglas \& Martin(2004)]{Douglas_Martin04}
Douglas, R. \& Martin, K. (2004).
\newblock Neural circuits of the neocortex.
\newblock Annual Review of Neuroscience, 27, 419--51.

\bibitem[Faisal et~al.(2008)Faisal, Selen, \& Wolpert]{Faisal_etal08}
Faisal, A., Selen, L., \& Wolpert, D. (2008).
\newblock Noise in the nervous system.
\newblock Nature reviews neuroscience, 9, 292--303.

\bibitem[Fries et~al.(2007)Fries, Nikoli{\'c}, \& Singer]{Fries_etal07}
Fries, P., Nikoli{\'c}, D., \& Singer, W. (2007).
\newblock The gamma cycle.
\newblock Trends in neurosciences, 30, 309--316.

\bibitem[Friston(2003)]{Friston03}
Friston, K. (2003).
\newblock Learning and inference in the brain.
\newblock Neural Networks, 16, 1325--1352.

\bibitem[Furber et~al.(2014)Furber, Galluppi, Temple, \& Plana]{Furber_etal14}
Furber, S., Galluppi, F., Temple, S., \& Plana, L. (2014).
\newblock The {SpiNNaker} project.
\newblock Proceedings of the IEEE, 102, 652--665.

\bibitem[Gewaltig \& Diesmann(2007)]{Gewaltig_Diesman07}
Gewaltig, M.-O. \& Diesmann, M. (2007).
\newblock Nest (neural simulation tool).
\newblock Scholarpedia, 2, 1430.

\bibitem[Goodfellow et~al.(2014)Goodfellow, Pouget-Abadie, Mirza, Xu,
  Warde-Farley, Ozair, Courville, \& Bengio]{Goodfellow14}
Goodfellow, I., Pouget-Abadie, J., Mirza, M., Xu, B., Warde-Farley, D., Ozair,
  S., Courville, A., \& Bengio, Y. (2014).
\newblock Generative adversarial nets.
\newblock In Advances in neural information processing systems, pp. 2672--2680.

\bibitem[Goodfellow et~al.(2013)Goodfellow, Warde-Farley, Mirza, Courville, \&
  Bengio]{Goodfellow_etal13}
Goodfellow, I., Warde-Farley, D., Mirza, M., Courville, A., \& Bengio, Y.
  (2013).
\newblock Maxout networks.
\newblock In ICML, 28, 1319--1327.

\bibitem[Gregory(1980)]{Gregory80}
Gregory, R. (1980).
\newblock Perceptions as hypotheses.
\newblock Philosophical Transactions of the Royal Society of London. B,
  Biological Sciences, 290, 181--197.

\bibitem[Gu et~al.(2015)Gu, Levine, Sutskever, \& Mnih]{Gu_etal14}
Gu, S., Levine, S., Sutskever, I., \& Mnih, A. (2015).
\newblock Muprop: Unbiased backpropagation for stochastic neural networks.
\newblock arXiv preprint arXiv:1511.05176.

\bibitem[Gumbel \& Lieblein(1954)]{Gumbel_Lieblein54}
Gumbel, E. \& Lieblein, J. (1954).
\newblock Statistical theory of extreme values and some practical applications:
  a series of lectures.

\bibitem[Hinton(2002)]{Hinton02}
Hinton, G. (2002).
\newblock Training products of experts by minimizing contrastive divergence.
\newblock Neural Computation, 14, 1771--1800.

\bibitem[Hunsberger \& Eliasmith(2015)]{Hunsberger_Eliasmith15}
Hunsberger, E. \& Eliasmith, C. (2015).
\newblock Spiking deep networks with {LIF} neurons.
\newblock arXiv preprint arXiv:1510.08829.

\bibitem[Jang et~al.(2017)Jang, Gu, \& Poole]{Jang_etal17}
Jang, E., Gu, S., \& Poole, B. (2017).
\newblock Categorical reparameterization with {Gumbel-softmax}.
\newblock Stat, 1050, 1.

\bibitem[Kingma \& Ba(2014)]{Kingma_Ba14}
Kingma, D. \& Ba, J. (2014).
\newblock Adam: A method for stochastic optimization.
\newblock arXiv preprint arXiv:1412.6980.

\bibitem[Kingma \& Welling(2013)]{Kingma_Welling13}
Kingma, D. \& Welling, M. (2013).
\newblock Auto-encoding variational {Bayes}.
\newblock arXiv preprint arXiv:1312.6114.

\bibitem[Larochelle \& Bengio(2008)]{Larochelle_Bengio08}
Larochelle, H. \& Bengio, Y. (2008).
\newblock Classification using discriminative restricted boltzmann machines.
\newblock In Proceedings of the 25th international conference on Machine
  learning, pp. 536--543. ACM.

\bibitem[Lee \& Mumford(2003)]{Lee_Mumford03}
Lee, T. \& Mumford, D. (2003).
\newblock Hierarchical {Bayesian} inference in the visual cortex.
\newblock Journal of the Optical Society of America. A, Optics, image science,
  and vision, 20, 1434--1448.

\bibitem[Lillicrap et~al.(2016)Lillicrap, Cownden, Tweed, \&
  Akerman]{Lillicrap_etal16}
Lillicrap, T., Cownden, D., Tweed, D., \& Akerman, C. (2016).
\newblock Random synaptic feedback weights support error backpropagation for
  deep learning.
\newblock Nature communications, 7.

\bibitem[Ma et~al.(2006)Ma, Beck, Latham, \& Pouget]{Ma_etal06}
Ma, W., Beck, J., Latham, P., \& Pouget, A. (2006).
\newblock Bayesian inference with probabilistic population codes.
\newblock Nature Neurosci, 9, 1432--1438.

\bibitem[Maddison et~al.(2016)Maddison, Mnih, \& Teh]{Maddison_etal16}
Maddison, C., Mnih, A., \& Teh, Y. (2016).
\newblock The concrete distribution: A continuous relaxation of discrete random
  variables.
\newblock arXiv preprint arXiv:1611.00712.

\bibitem[Mostafa et~al.(2015)Mostafa, M\"uller, \& Indiveri]{Mostafa_etal15}
Mostafa, H., M\"uller, L.~K., \& Indiveri, G. (2015).
\newblock Rhythmic inhibition allows neural networks to search for maximally
  consistent states.
\newblock Neural Computation, 27, 2510--2547.

\bibitem[Mostafa et~al.(2017)Mostafa, Ramesh, \& Cauwenberghs]{Mostafa_etal17a}
Mostafa, H., Ramesh, V., \& Cauwenberghs, G. (2017).
\newblock Deep supervised learning using local errors.
\newblock arXiv preprint arXiv:1711.06756.

\bibitem[Neftci et~al.(2014)Neftci, Das, Pedroni, Kreutz-Delgado, \&
  Cauwenberghs]{Neftci_etal14}
Neftci, E., Das, S., Pedroni, B., Kreutz-Delgado, K., \& Cauwenberghs, G.
  (2014).
\newblock Event-driven contrastive divergence for spiking neuromorphic systems.
\newblock Frontiers in Neuroscience, 7.

\bibitem[Neftci et~al.(2016)Neftci, Pedroni, Joshi, Al-Shedivat, \&
  Cauwenberghs]{Neftci_etal16}
Neftci, E., Pedroni, B., Joshi, S., Al-Shedivat, M., \& Cauwenberghs, G.
  (2016).
\newblock Stochastic synapses enable efficient brain-inspired learning
  machines.
\newblock Frontiers in neuroscience, 10.

\bibitem[Nessler et~al.(2013)Nessler, Pfeiffer, \& Maass]{Nessler_etal13}
Nessler, B., Pfeiffer, M., \& Maass, W. (2013).
\newblock Bayesian computation emerges in generic cortical microcircuits
  through spike-timing-dependent plasticity.
\newblock PLoS computational biology, 9, e1003037.

\bibitem[N{\o}kland(2016)]{Nokland_etal16}
N{\o}kland, A. (2016).
\newblock Direct feedback alignment provides learning in deep neural networks.
\newblock In Advances in Neural Information Processing Systems, pp. 1037--1045.

\bibitem[O'Connor et~al.(2013)O'Connor, Neil, Liu, Delbruck, \&
  Pfeiffer]{OConnor_etal13}
O'Connor, P., Neil, D., Liu, S.-C., Delbruck, T., \& Pfeiffer, M. (2013).
\newblock Real-time classification and sensor fusion with a spiking deep belief
  network.
\newblock Frontiers in Neuroscience, 7.

\bibitem[Panda \& Roy(2016)]{Panda_Roy16}
Panda, P. \& Roy, K. (2016).
\newblock Unsupervised regenerative learning of hierarchical features in
  spiking deep networks for object recognition.
\newblock In Neural Networks (IJCNN), 2016 International Joint Conference on,
  pp. 299--306. IEEE.

\bibitem[Park et~al.(2014)Park, Ha, Yu, Neftci, \& Cauwenberghs]{Park_etal14}
Park, J., Ha, S., Yu, T., Neftci, E., \& Cauwenberghs, G. (2014).
\newblock A 65k-neuron 73-{Mevents/s} 22-{pJ}/event asynchronous
  micro-pipelined integrate-and-fire array transceiver.
\newblock In Biomedical Circuits and Systems Conference (BioCAS), 2014 IEEE,
  pp. 675--678. IEEE.

\bibitem[Pecevski et~al.(2011)Pecevski, Buesing, \& Maass]{Pecevski_etal11}
Pecevski, D., Buesing, L., \& Maass, W. (2011).
\newblock Probabilistic inference in general graphical models through sampling
  in stochastic networks of spiking neurons.
\newblock PLoS computational biology, 7, e1002294.

\bibitem[Qiao et~al.(2015)Qiao, Mostafa, Corradi, Osswald, Stefanini,
  Sumislawska, \& Indiveri]{Qiao_etal15}
Qiao, N., Mostafa, H., Corradi, F., Osswald, M., Stefanini, F., Sumislawska,
  D., \& Indiveri, G. (2015).
\newblock A re-configurable on-line learning spiking neuromorphic processor
  comprising 256 neurons and 128k synapses.
\newblock Frontiers in Neuroscience, 9.

\bibitem[Raiko et~al.(2014)Raiko, Berglund, Alain, \& Dinh]{Raiko_etal14}
Raiko, T., Berglund, M., Alain, G., \& Dinh, L. (2014).
\newblock Techniques for learning binary stochastic feedforward neural
  networks.
\newblock Stat, 1050, 11.

\bibitem[Rasmus et~al.(2015)Rasmus, Berglund, Honkala, Valpola, \&
  Raiko]{Rasmus_etal15}
Rasmus, A., Berglund, M., Honkala, M., Valpola, H., \& Raiko, T. (2015).
\newblock Semi-supervised learning with ladder networks.
\newblock In Advances in Neural Information Processing Systems, pp. 3546--3554.

\bibitem[Rezende et~al.(2014)Rezende, Mohamed, \& Wierstra]{Rezende_etal14}
Rezende, D., Mohamed, S., \& Wierstra, D. (2014).
\newblock Stochastic backpropagation and approximate inference in deep
  generative models.
\newblock In ICML, pp. 1278--1286.

\bibitem[Rotter \& Aertsen(1998)]{Rotter_Aertsen98}
Rotter, S. \& Aertsen, A. (1998).
\newblock Accurate spike synchronization in cortex.
\newblock Zeitschrift f{\"u}r Naturforschung C, 53, 686--690.

\bibitem[Ruiz et~al.(2016)Ruiz, Titsias, \& Blei]{Ruiz_etal16}
Ruiz, F., Titsias, M., \& Blei, D. (2016).
\newblock The generalized reparameterization gradient.
\newblock In Advances in Neural Information Processing Systems, pp. 460--468.

\bibitem[Scellier \& Bengio(2017)]{Scellier_Bengio17}
Scellier, B. \& Bengio, Y. (2017).
\newblock Equilibrium propagation: Bridging the gap between energy-based models
  and backpropagation.
\newblock Frontiers in computational neuroscience, 11.

\bibitem[S{\o}nderby et~al.(2016)S{\o}nderby, Raiko, Maal{\o}e, S{\o}nderby, \&
  Winther]{Sonderby_etal16}
S{\o}nderby, C., Raiko, T., Maal{\o}e, L., S{\o}nderby, S., \& Winther, O.
  (2016).
\newblock Ladder variational autoencoders.
\newblock In Advances in Neural Information Processing Systems, pp. 3738--3746.

\bibitem[Srivastava et~al.(2014)Srivastava, Masci, Gomez, \&
  Schmidhuber]{Srivastava_etal14a}
Srivastava, R., Masci, J., Gomez, F., \& Schmidhuber, J. (2014).
\newblock Understanding locally competitive networks.
\newblock arXiv preprint arXiv:1410.1165.

\bibitem[Srivastava et~al.(2013)Srivastava, Masci, Kazerounian, Gomez, \&
  Schmidhuber]{Srivastava13}
Srivastava, R., Masci, J., Kazerounian, S., Gomez, F., \& Schmidhuber, J.
  (2013).
\newblock Compete to compute.
\newblock In Advances in neural information processing systems, pp. 2310--2318.

\bibitem[Stratford et~al.(1996)Stratford, Tarczy-Hornoch, Martin, Bannister, \&
  Jack]{Stratford_etal96}
Stratford, K., Tarczy-Hornoch, K., Martin, K., Bannister, N., \& Jack, J.
  (1996).
\newblock Excitatory synaptic inputs to spiny stellate cells in cat visual
  cortex.
\newblock Nature, 382, 258.

\bibitem[Valpola(2015)]{Valpola15}
Valpola, H. (2015).
\newblock From neural {PCA} to deep unsupervised learning.
\newblock Adv. in Independent Component Analysis and Learning Machines, pp.
  143--171.

\bibitem[Vincent et~al.(2008)Vincent, Larochelle, Bengio, \&
  Manzagol]{Vincent_etal08}
Vincent, P., Larochelle, H., Bengio, Y., \& Manzagol, P.-A. (2008).
\newblock Extracting and composing robust features with denoising autoencoders.
\newblock In Proceedings of the 25th international conference on Machine
  learning, pp. 1096--1103. ACM.

\bibitem[Volgushev et~al.(1995)Volgushev, Voronin, Chistiakova, Artola, \&
  Singer]{Volgushev_etal95}
Volgushev, M., Voronin, L., Chistiakova, M., Artola, A., \& Singer, W. (1995).
\newblock All-or-none excitatory postsynaptic potentials in the rat visual
  cortex.
\newblock European Journal of Neuroscience, 7, 1751--1760.

\bibitem[Williams(1992)]{Williams92}
Williams, R. (1992).
\newblock Simple statistical gradient-following algorithms for connectionist
  reinforcement learning.
\newblock Machine learning, 8, 229--256.

\bibitem[Xie \& Seung(2003)]{Seung_etal03}
Xie, X. \& Seung, S. (2003).
\newblock Equivalence of backpropagation and contrastive hebbian learning in a
  layered network.
\newblock Neural computation, 15, 441--454.

\end{thebibliography}

\end{document}